\documentclass[11pt]{article}

\usepackage[preprint]{acl}
\usepackage{times}
\usepackage{latexsym}
\usepackage[T1]{fontenc}
\usepackage[utf8]{inputenc}
\usepackage{microtype}
\usepackage{graphicx}
\usepackage{xspace}
\usepackage{amsmath, amssymb}
\usepackage{enumitem}
\usepackage{array, booktabs}
\usepackage[table]{xcolor}
\usepackage{multirow}
\usepackage{arydshln}
\usepackage{makecell}
\usepackage{cuted}

\newtheorem{theorem}{Theorem}

\newcommand{\our}{\textsc{PSOT}\xspace}
\newcommand{\ourone}{Peak Suppression Orthogonal Transformation\xspace}
\newcommand{\AT}{ASOT\xspace}
\newcommand{\ours}{\textsc{InfoQuant}\xspace}
\newcommand{\ourse}{\textsc{InfoQuant*}\xspace}

\title{\ours: Shaping Activation Distributions for Low-Bit \\ LLM Quantization}

\author{
  \textbf{Ke Li\textsuperscript{1}},
  \textbf{Dong An\textsuperscript{2}},
  \textbf{Xiaoling Zang\textsuperscript{2}},
  \textbf{Can Ye\textsuperscript{2}},
  \textbf{Liang Xie\textsuperscript{3}},
\\
  \textbf{Qibo Qiu\textsuperscript{5}},
  \textbf{Chen Shen\textsuperscript{4}},
  \textbf{Xiaofei He\textsuperscript{5}},
  \textbf{Wenxiao Wang\textsuperscript{1,6,*}}
\\
\\
  \textsuperscript{1}School of Software Technology, Zhejiang University \quad\textsuperscript{2}Ant Group
\\
  \textsuperscript{3}Zhejiang University of Technology  \quad \textsuperscript{4}Alibaba Group
\\
  \textsuperscript{5}State Key Lab of CAD\&CG, Zhejiang University \quad \textsuperscript{6}Darwinmind AI
\\
  \small{\texttt{\char`\{like2248,wenxiaowang\char`\}@zju.edu.cn}}
}

\hypersetup{
  pdftitle={InfoQuant: Shaping Activation Distributions for Low-Bit LLM Quantization},
  pdfauthor={Ke Li; Dong An; Xiaoling Zang; Can Ye; Liang Xie; Qibo Qiu; Chen Shen; Xiaofei He; Wenxiao Wang},
  pdfsubject={Post-training quantization of large language models through activation-distribution shaping},
  pdfkeywords={large language models, post-training quantization, activation quantization, low-bit inference, orthogonal transformation, information-theoretic analysis}
}

\begin{document}
\maketitle
\begingroup
\renewcommand{\thefootnote}{*}
\footnotetext{Corresponding author.}
\endgroup
\begin{abstract}
Low-bit activation quantization remains a major bottleneck in efficient large language model (LLM) deployment. The difficulty is not only that activations contain outliers, but that their distributions are often poorly matched to a low-bit uniform quantizer. Existing post-training quantization (PTQ) methods suppress peaks, balance channels, or minimize reconstruction error, yet they rarely specify what activation distribution is actually easy to discretize. As a result, activations may appear numerically smoother while still incurring large quantization error because the quantization range remains wide or most values collapse into a few levels near the mean.
We recast activation transformation as quantizer-facing distribution design and analyze quantization error from an information-theoretic perspective. Our analysis shows that quantization-friendly activations should jointly have a smaller numerical range and sufficient dispersion within that range.
Guided by this analysis, we propose \ours, a train-free method that employs \ourone (\our) to shape activations into more quantization-friendly distributions. We further introduce adaptive outlier-token selection to improve the robustness of \our during optimization.
Across multiple LLM families, \ours consistently outperforms prior PTQ and end-to-end training baselines. Under W4A4KV4, it preserves 97\% of floating-point accuracy on average.\footnote{Code is available at: \href{https://github.com/LLIKKE/InfoQuant}{github.com/LLIKKE/InfoQuant}.}
\end{abstract}

\section{Introduction}
Post-training quantization (PTQ) is one of the most practical ways to reduce the memory and compute cost of large language model (LLM) inference. Its main challenge, however, lies in low-bit activation quantization. Unlike weights, LLM activations often contain a small number of dominant coordinates that enlarge the quantization range and force round-to-nearest quantization to map many normal values to the same few levels. This mismatch becomes particularly severe in 4-bit settings, where limited quantization levels leave little room to preserve both rare extremes and dense central values.

\begin{figure*}[t]
	\centering
	\includegraphics[width=0.95\textwidth]{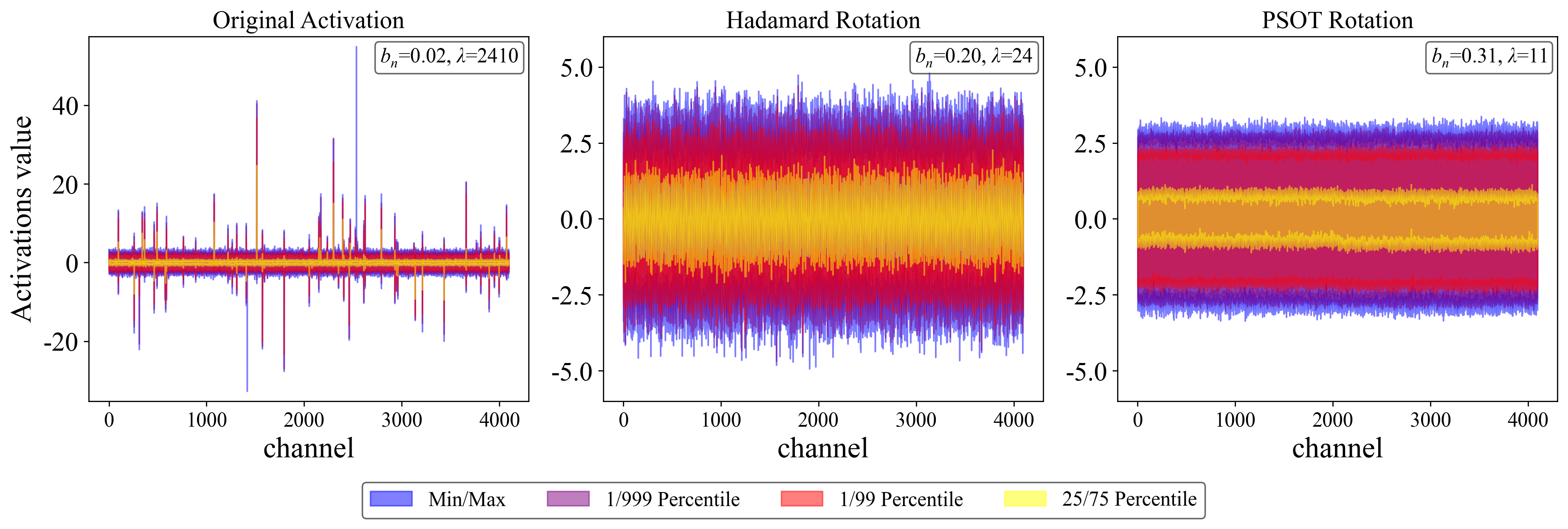}
	\caption{Activation distributions of the \texttt{LLaMA-2 7B, layer4 q/k/v\_proj} input under three transformations: the original activations (left), Hadamard rotation (center), and the learned rotation from \our (right). Compared with the original and Hadamard-rotated activations, \our produces a more quantization-friendly distribution with a narrower numerical range and larger normalized dispersion. Here, $b_n$ denotes the standard deviation after infinity-norm normalization, and a smaller $\lambda=\bar{s}/b_n$ indicates a lower normalized quantization-error bound, where $\bar{s}$ is the range-normalized quantization step size.}
	\label{oho}
\end{figure*}

Recent PTQ methods increasingly address this problem through activation transformations before quantization. SmoothQuant~\cite{xiao2023smoothquant} migrates activation difficulty into weights through diagonal scaling, while QuaRot~\cite{ashkboos2024quarot} and SpinQuant~\cite{liu2025spinquant} use orthogonal rotations to redistribute activation energy; other methods further introduce more flexible affine transformations or reconstruction objectives~\cite{MaLZLX0W0J24,sun2025flatquantflatnessmattersllm}. Although these methods differ in form, they share the same practical role: they change the activation distribution seen by the quantizer. Yet most of them are motivated by suppressing outliers, balancing channels, or reducing reconstruction error, rather than by defining what transformed distribution a low-bit quantizer can represent well. Consequently, as illustrated in Figure~\ref{oho}, they may reduce visible peaks without fully improving discretizability, or preserve small numerical error while still collapsing distributional resolution. The central question, therefore, is not only how to transform activations, but what transformed activation distribution is actually quantization-friendly.

We address this gap by recasting activation transformation as quantizer-facing distribution design. From an information-theoretic perspective, we analyze how quantization error depends on the activation distribution after transformation. Our theoretical and empirical study shows that lower quantization error is associated with two complementary properties: a smaller numerical range and greater dispersion within that range. This result reframes the role of activation transformation. Rather than merely suppressing outliers or minimizing heuristic reconstruction losses, a PTQ method should explicitly shape activations into distributions that are easier for a low-bit quantizer to preserve. Since LLM activations are typically bell-shaped and often contain outliers~\cite{liu2025trainingfree}, they are naturally misaligned with this target, which explains why low-bit activation quantization remains difficult.

Guided by this principle, we introduce \textsc{InfoQuant}, a train-free PTQ method that learns orthogonal transformations to produce more quantization-friendly activation distributions. Its core component is \ourone (\our), which applies an activation-wise peak suppression objective to reduce the numerical range while increasing normalized dispersion. We further introduce adaptive outlier-token selection to improve optimization robustness, and learn activation clipping parameters to refine the final quantization range after the distribution has been reshaped. Although \ours is designed around activation optimization, it remains compatible with standard weight quantization pipelines. Overall, our contributions can be summarized as follows:
\begin{itemize}[leftmargin=2em]
	\item We introduce an information-theoretic framework for understanding activation quantization error and show, both theoretically and empirically, that quantization-friendly distributions should have a smaller numerical range and greater dispersion.
	\item We introduce \textsc{InfoQuant}, a hardware-efficient and train-free PTQ method centered on learned orthogonal activation shaping, together with adaptive outlier-token selection and learnable activation clipping for robust calibration.
	\item We demonstrate that activation-distribution optimization yields strong empirical and practical gains. In the W4A4KV4 setting, LLaMA-2 (7B, 13B, 70B) and LLaMA-3 (8B, 70B) retain an average of $97\%$ of their original performance, and the 70B model can be quantized using only $24$GB of GPU memory.
\end{itemize}

\section{Related Work}
\paragraph{Post Training Quantization for LLMs.}
PTQ is an efficient and widely used approach for compressing LLMs. Due to the flatness and uniform distribution of LLM weights, weight-only quantization typically results in minimal performance degradation. GPTQ~\cite{frantar2023optq} uses Hessian-based error compensation to enable high compression with low accuracy loss. AWQ~\cite{lin2024awq} and OWQ~\cite{lee2024owq} further improve performance by mitigating the effects of activation outliers. QuIP~\citep{chee2023quip} and QuIP\#~\citep{tseng2024quip} apply random Hadamard transforms for incoherent processing and employ vector quantization on weights, achieving better performance.
In contrast, activation quantization remains more challenging due to the presence of rare but extreme outliers~\citep{wei2023outlier,xiao2023smoothquant}, which can disproportionately affect accuracy.
\paragraph{Transformation-based Methods.}
These methods more effectively redistribute activation outliers across channels. Channel scaling~\citep{xiao2023smoothquant} shifts part of this burden to weights, OmniQuant~\citep{shao2024omniquant} and LRQuant~\citep{zhao2024lrquant} optimize scaling parameters via MSE minimization. However, recent work~\citep{yi2025rotated} shows channel scaling alone fails under 4-bit settings, leading to notable degradation.
AffineQuant~\citep{MaLZLX0W0J24} learns affine transformations to precondition activations. However, due to the significant overhead of full-size matrix multiplication, AffineQuant can only apply affine transformations to a small fraction of linear layers. FlatQuant~\citep{sun2025flatquantflatnessmattersllm} reduces this cost via Kronecker decomposition, applying affine transformation to every linear layer. Leveraging computational invariance~\citep{ashkboos2024slicegpt}, orthogonal transforms can be applied to weights and between-block activations without extra inference overhead. QuaRot~\citep{ashkboos2024quarot} uses randomized Hadamard transforms to remove outliers. DuQuant~\citep{lin2024duquant} redistributes outliers through online rotations, which require specialized kernels for efficient low-bit inference. SpinQuant~\citep{liu2025spinquant} further optimizes learnable orthogonal matrices on the Stiefel manifold with task loss (e.g., cross-entropy) to find stable transformations. OSTQuant~\cite{hu2025ostquant} combines channel scaling with orthogonal transforms and uses end-to-end distillation from original outputs to boost quantization. Kurtail~\citep{akhondzadeh2025kurtail} facilitates quantization by controlling the kurtosis to make the distribution more uniform. BASE-Q~\citep{he2025base} introduces an additional bias term to balance the mean values of different channels after rotation.

\section{Motivation}
\subsection{Quantization Preliminaries}
Quantization maps high-precision values to a set of discrete levels. The process is detailed as follows:
\begin{equation}
	\mathcal{Q}(\mathbf{X}) = \text{clamp}\left( \left\lfloor \frac{\mathbf{X}}{s} \right\rceil + z,\ 0,\ 2^N - 1 \right)
\end{equation}
Here, quantization step size is denoted by $s = \frac{\mathbf{X}_{\max} - \mathbf{X}_{\min}}{2^N - 1}$, and $z = -\left\lfloor \frac{\mathbf{X}_{\min}}{s} \right\rceil$ is the corresponding zero-point, $\left\lfloor \cdot \right\rceil$ denotes the rounding operation, and $N$ represents the target bit-width. Given a floating-point tensor $\mathbf{X}$, the quantization function $\mathcal{Q}(\cdot)$ produces its integer-valued representation.
Quantization error primarily arises from the rounding operation, which collapses all values within a single interval of size $s$ into the same discrete level.
\subsection{A Distributional View of Quantization Error}\label{sec:KL}
Recent PTQ methods often optimize activation transformations with MSE-based objectives~\citep{shao2024omniquant,zhao2024lrquant,sun2025flatquantflatnessmattersllm}. While MSE is a useful measure of numerical distortion, it does not fully capture the distributional mismatch introduced by low-bit quantization.

This limitation is especially important for activation quantization under round-to-nearest (RTN), where many values may incur only small pointwise errors yet still be mapped to a small number of discrete levels. In such cases, the quantized activations can remain close in value to the original ones while losing substantial distributional resolution, which is not well reflected by MSE alone.
\begin{figure}[t]
	\centering
	\includegraphics[width=0.95\linewidth]{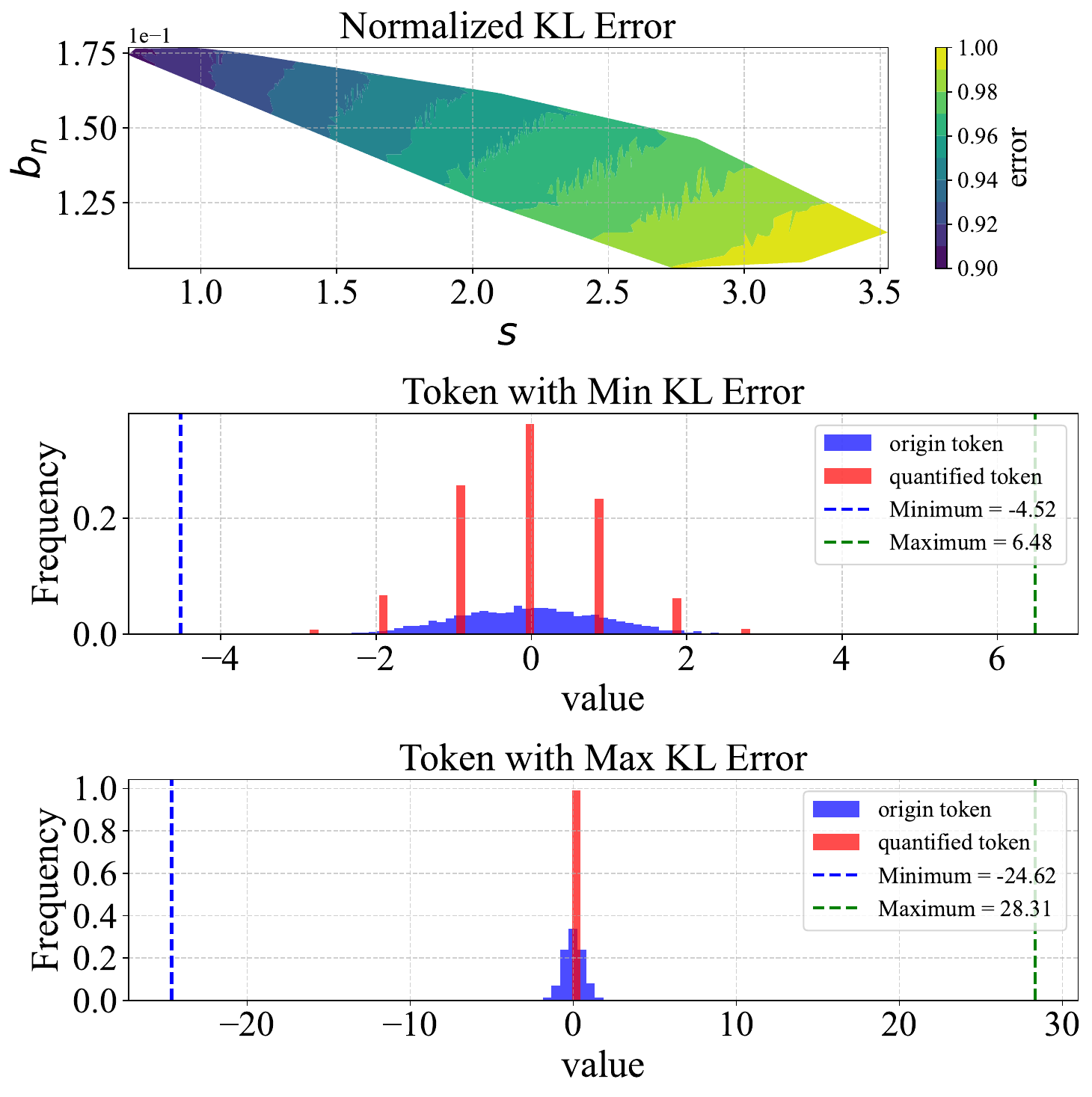}
	\caption{Distributional effect of quantizing the \texttt{LLaMA-2 7B, layer4 q/k/v\_proj} input.
		Top: KL divergence between activation histograms before and after quantization, evaluated over different quantization step sizes $s$ and dispersion values $b_n$ with 15,000 histogram bins.
		Center/Bottom: activation histograms before and after quantization for low-error and high-error cases, respectively. Low-bit quantization is most destructive when a wide range and low normalized dispersion force dense activation values into too few discrete levels.}
	\label{mse_kl}
\end{figure}
Prior work~\citep{liu2025trainingfree} has shown that activation distributions in LLMs are typically bell-shaped (e.g., Gaussian or Laplace). When such activations are quantized with a low-bit uniform RTN quantizer, a few large values can determine the quantization range, forcing most normal values to collapse toward levels near the mean (Figure~\ref{mse_kl}, bottom). The resulting error is therefore not only a matter of local rounding distortion, but also of how poorly the available quantization levels match the underlying activation distribution.

Non-uniform quantizers can in principle better adapt to dense regions; however, they usually introduce additional hardware complexity and are less attractive in practical low-bit deployment. These observations motivate an analytical metric that reflects both numerical deviation and quantization-induced distribution shift.
To address this, we use a smoothed KL divergence as an analytical lens for the distributional distortion caused by low-bit quantization. Let $\mathbf{x}$ denote an activation token, and let each entry be denoted by a scalar $x \in \mathbf{x}$ with distribution $P(x)$. We consider a centered finite-bit clamped quantizer
\begin{equation}
	\hat{x}=Q_{s,c}(x)=\operatorname{clip}\!\left(s\left\lfloor \frac{x}{s}\right\rceil,\,-c,\,c\right),
\end{equation}
where $s$ is the quantization step size and $c$ is the clipping scale. Directly comparing $P(x)$ with the quantized distribution is ill-posed as a density-to-density KL, because quantization turns a continuous density into probability masses on discrete centroids. We therefore spread each centroid into a narrow continuous kernel. Specifically, let $q_i$ denote a quantization centroid, let $I_i=\{x\mid Q_{s,c}(x)=q_i\}$ be its quantization cell, and define the corresponding probability mass $p_i=\int_{I_i}P(x)\,dx$. The quantized distribution is relaxed as
\begin{equation}
	Q_\theta^{(c)}(x)\approx \sum_i p_i \cdot \delta(x-q_i;\theta),
\end{equation}
where $\delta(x;\theta)=\frac{1}{2\theta}\exp\!\left(-\frac{|x|}{\theta}\right)$ is a Laplace kernel that approaches a Dirac delta as $\theta\to 0$. Under the standard separation assumption $\theta\ll s$, the smoothed KL objective admits the approximation:
\vspace{-2pt}
\begin{align}
	&D_{\mathrm{KL}}\left(P \,\|\, Q_\theta^{(c)}\right)
	= \int P(x)\, \log\!\left(\frac{P(x)}{Q_\theta^{(c)}(x)} \right) dx \nonumber \\
	&\approx -H(P) + H(\{p_i\}) + \log(2\theta) + \frac{1}{\theta}\, \mathcal{E}_{\mathrm{clip}},
\end{align}
where $H(P)$ denotes the entropy of the original distribution, $H(\{p_i\})$ is the entropy of the quantized probability masses, and $\mathcal{E}_{\mathrm{clip}}=\mathbb{E}|x-Q_{s,c}(x)|$ is the expected absolute error of the finite-bit clamped quantizer. This decomposition exposes the key mechanism: after smoothing the discrete outputs, the distributional KL surrogate contains a direct quantization-error term. Thus, a transformation that makes activations easier to quantize should not only suppress extreme values, but also reduce the normalized error induced by the finite set of quantization levels.

Since activations typically have bounded yet varying ranges, we normalize the clamped quantization error by the standard deviation $\sigma$:
\begin{equation}
\begin{split}
	\mathcal{E}_{\mathrm{clip}}'
	&= \frac{1}{\sigma}\,\mathbb{E}|x-Q_{s,c}(x)| \\
	&\le \frac{s}{2\sigma}
	+ \frac{1}{\sigma}\,\mathbb{E}\!\left[(|x|-c)_+\right].
\end{split}
\end{equation}
The two terms reveal the trade-off that ordinary outlier suppression does not fully describe. The first term is the in-range rounding error, which decreases when the step size is small relative to the activation spread. The second term is the clipping-tail error, which measures the mass left outside the finite range. Let $\lambda=s/\sigma$ and $\kappa=c/\sigma$. Then
\begin{equation}
	\mathcal{E}_{\mathrm{clip}}' \le \frac{\lambda}{2} + \tau_P(\kappa),
\end{equation}
where $\tau_P(\kappa)=\mathbb{E}\!\left[(|Y|-\kappa)_+\right]$ for the normalized variable $Y=x/\sigma$. For common bell-shaped activation distributions such as Gaussian and Laplace, $\tau_P$ admits closed forms, and for a $B$-bit quantizer with $c=Ms$ and $M=2^{B-1}-1$, the resulting bound decreases as $\kappa$ decreases in the tail-controlled regime relevant to calibrated PTQ.\footnote{Proofs and the Gaussian/Laplace closed forms can be found in Appendix~\ref{sec:appx_proofs}.} Since $\kappa=c/\sigma=1/b_n$ and $\lambda=s/\sigma=\bar{s}/b_n$ with $b_n=\sigma/c$, this analysis turns the vague goal of ``making activations smoother'' into a concrete distributional target: reduce the clipped numerical range while keeping the normalized activation values well dispersed inside that range. Empirically, as shown in the top of Figure~\ref{mse_kl}, KL divergence decreases consistently with smaller $s$ and larger $b_n$. Together, the analysis and observation suggest a simple design principle for low-bit activation quantization: a good transformation should compress the effective range and spread useful activation mass across more available quantization levels.

\section{Method}

\ours is a train-free PTQ framework that reshapes activations into distributions better matched to low-bit quantizers. It consists of three components: \ourone (\our) learns an orthogonal activation transformation, adaptive outlier-token selection (\AT) emphasizes informative calibration tokens, and learnable activation clipping (LAC) refines the final quantization interval.

\subsection{\ourone}
\our learns a quantizer-facing orthogonal transformation by directly penalizing peak-dominated activation tokens. For each target activation stream, we optimize a block-diagonal orthogonal rotation on calibration activations and initialize it from a Hadamard transform, preserving the efficient rotation-based deployment path used by existing PTQ systems. Let $\mathbf{x}\in\mathbb{R}^{d}$ denote one activation token, and let $\mathbf{R}\in\mathbb{R}^{d\times d}$ be a learnable orthogonal matrix with $\mathbf{R}^{\top}\mathbf{R}=\mathbf{I}$. We remove the coordinate-wise mean after rotation using the centering projector $\mathbf{P}_{\perp}=\mathbf{I}-\frac{1}{d}\mathbf{1}\mathbf{1}^{\top}$, and define $\boldsymbol{\pi}_T(\mathbf{y})=\operatorname{softmax}(|\mathbf{y}|/T)$:
\begin{align}
	\mathbf{y}(\mathbf{x};\mathbf{R}) &= \mathbf{x}\mathbf{R}\mathbf{P}_{\perp}, \nonumber \\
	\ell_{\mathrm{ps}}(\mathbf{x};\mathbf{R})
	&=
	\left\|
	\boldsymbol{\pi}_T(\mathbf{y})
	\odot \mathbf{y}
	\right\|_2 ,
	\label{eq:ommin}
\end{align}
where $T$ is the temperature and $\odot$ denotes element-wise multiplication. The softmax weights concentrate the objective on high-magnitude coordinates, so minimizing $\ell_{\mathrm{ps}}$ suppresses the coordinates that dominate the quantization range.

This peak-suppression objective promotes the two distributional properties identified by the KL analysis. Since $\mathbf{R}$ is orthogonal, the rotation preserves token energy before centering. Reducing the largest centered coordinate therefore pushes the remaining energy to spread across more dimensions, which lowers the clipping scale and increases the max-normalized dispersion $b_n$. As illustrated in Figure~\ref{oho}, the learned rotation yields a narrower range and a more dispersed normalized distribution than both the original activations and a fixed Hadamard rotation.\footnote{Appendix~\ref{sec:appx_proofs} gives the corresponding formal analysis.}

\subsection{Adaptive Outlier-Token Selection}
\our obtains the strongest learning signal from tokens with quantization-sensitive outlier structure. Uniformly optimizing all calibration tokens can dilute this signal because many tokens have weak or noisy peaks. We therefore use \AT to select reliable outlier tokens and reweight them during rotation optimization.

Let $\mathbf{X}^{(r)}\in\mathbb{R}^{n_r\times d}$ be the activation matrix of the $r$-th calibration sample, where $\mathbf{x}^{(r)}_i$ is its $i$-th token. For a threshold coefficient $k$, we define the selected token-index set as
\begin{align}
	o_i^{(r)}
	&=
	\left\|
	\frac{\mathbf{x}^{(r)}_i-\mu\mathbf{1}}{\sigma}
	\right\|_{\infty},
	\nonumber\\
	\mathcal{T}^{(r)}(k)
	&=
	\left\{
	i \mid o_i^{(r)} > k
	\right\},
	\label{eq:token_set}
\end{align}
where $\mu$ and $\sigma$ are estimated from the corresponding calibration activations. This criterion keeps tokens that contain at least one statistically extreme coordinate.

The threshold should select sparse outlier tokens without overfitting to fixed sequence positions. Fixed-position peaks can reflect prompt layout or calibration artifacts, whereas sample-dependent outliers provide a more useful signal for learning a rotation that generalizes across inputs. Following the observation that activation outliers are more closely tied to token identity than absolute sequence position~\citep{liu2024intactkv,chen2025prefixquanteliminatingoutliersprefixed}, we measure whether the selected positions vary across $m$ calibration samples:
\begin{align}
	\eta_m(k)
	=
	1 -
	\frac{\frac{1}{m}\sum_{r=1}^{m}|\mathcal{T}^{(r)}(k)|}
	{\left|\bigcup_{r=1}^{m}\mathcal{T}^{(r)}(k)\right|}.
	\label{eq:eta}
\end{align}
where a larger $\eta_m(k)$ indicates that the selected outlier positions are less tied to fixed sequence locations. We choose the smallest threshold at which the inconsistency curve stabilizes while the selected tokens remain sparse:
\begin{align}
	k^{\star}
	=
	\min
	\left\{
	k\in\mathcal{K}
	\,\middle|\,
	\begin{aligned}
		&|\nabla\eta_m(k)| < \delta,\\
		&\frac{1}{m}\sum_{r=1}^{m}|\mathcal{T}^{(r)}(k)|
		< \tau |\mathcal{I}|
	\end{aligned}
	\right\},
	\label{selek}
\end{align}
where $\mathcal{K}$ is the ordered threshold grid, $\mathcal{I}$ is the token-index universe, $\delta$ controls the stabilization tolerance, and $\tau$ limits the selected-token ratio. With the selected sets $\mathcal{T}^{(r)}(k^\star)$, the final rotation objective is
\begin{align}
	\min_{\mathbf{R}^{\top}\mathbf{R}=\mathbf{I}}
	&\sum_{r=1}^{m}\sum_{i\in\mathcal{I}}
	w_i^{(r)}\,\ell_{\mathrm{ps}}(\mathbf{x}^{(r)}_i;\mathbf{R}),
	\nonumber\\
	w_i^{(r)}
	&=
	\begin{cases}
		\gamma, & i\in\mathcal{T}^{(r)}(k^\star),\\
		1, & \text{otherwise}.
	\end{cases}
	\label{eq:gamma}
\end{align}
where $\gamma>1$ emphasizes outlier tokens while retaining normal tokens as regularizing calibration samples. All \AT hyperparameters are selected by grid search on the calibration set, and the effect of $\gamma$ is studied in Table~\ref{tab:ablation_gamma}.
\vspace{-8pt}
\begin{figure}[htp]
	\centering \includegraphics[width=1\linewidth]{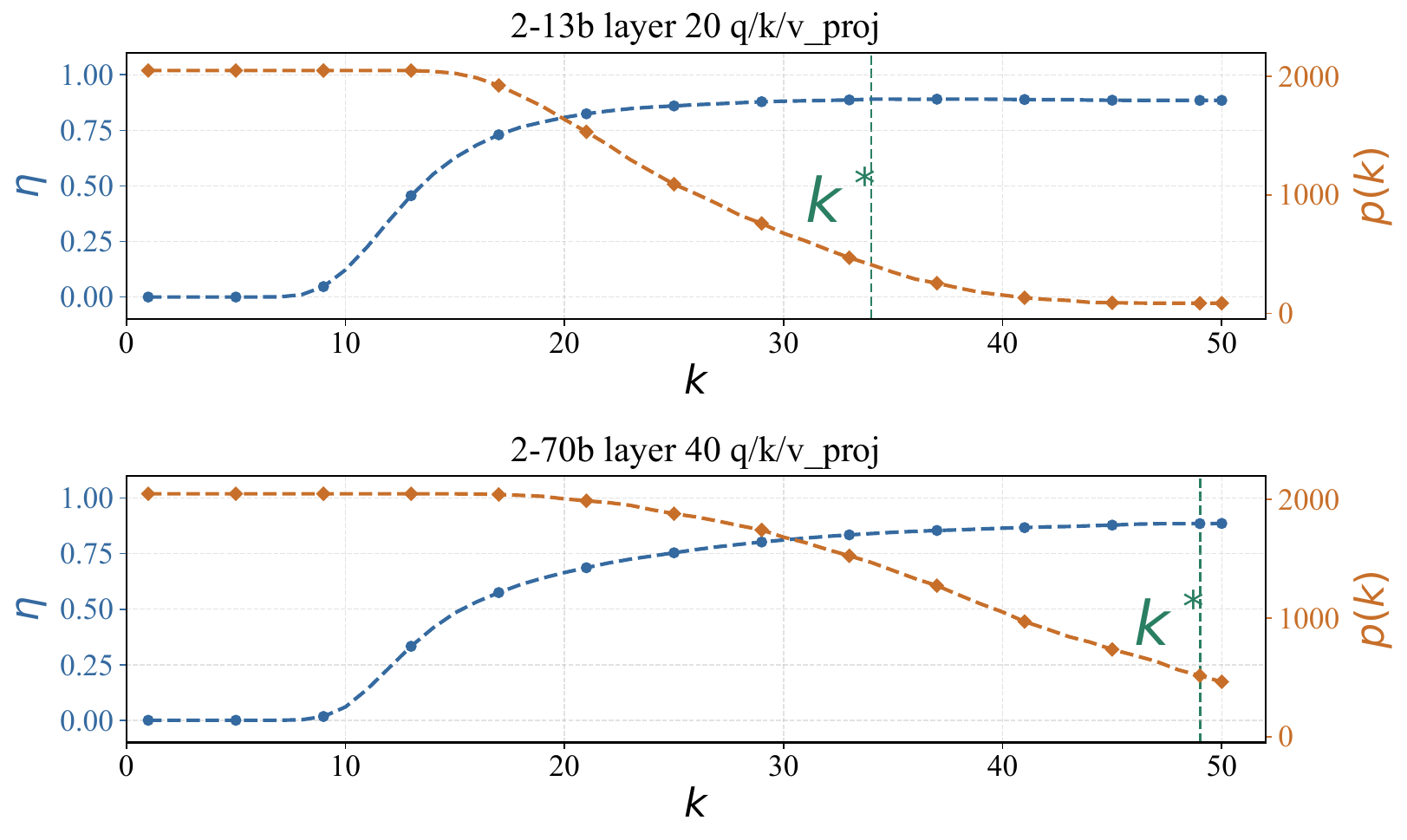}
	\caption{Adaptive threshold selection for \AT. The positional inconsistency $\eta_{10}(k)$ increases with $k$ and then plateaus, while the average number of selected outlier tokens $p(k)=\frac{1}{m}\sum_{r=1}^m|\mathcal{T}^{(r)}(k)|$ decreases. We choose $k^\star$ when $\eta_{10}(k)$ stabilizes and the selected token ratio satisfies the sparsity constraint in Eq.~\eqref{selek}.}
	\label{k}
\end{figure}

\subsection{Learnable Activation Clipping}
After \our reshapes the activation distribution, we apply learnable activation clipping as a final calibration step. Following prior work~\citep{zhao2024lrquant,sun2025flatquantflatnessmattersllm}, two bounded parameters $\alpha$ and $\beta$ refine the activation step size:
\begin{align}
	s(\alpha,\beta)
	=
	\frac{\alpha X_{\max}-\beta X_{\min}}{2^N-1},
	\label{eq:lac_step}
\end{align}
where $N$ is the bit-width, and $X_{\max}$ and $X_{\min}$ are the observed activation bounds. The clipping parameters are optimized by matching the quantized block output to the full-precision block output:
\begin{align}
	\min_{\alpha,\beta}
	\left\|
	\mathcal{F}_{l}(\mathbf{X}_{l})
	-
	\widehat{\mathcal{F}}_{l}(\mathbf{X}_{l};\alpha,\beta)
	\right\|_2^2,
	\label{eq:lac_obj}
\end{align}
where $\mathcal{F}_{l}$ denotes the full-precision Transformer block and $\widehat{\mathcal{F}}_{l}$ denotes the same block evaluated with activation quantization under Eq.~\eqref{eq:lac_step}.

\section{Experiments}\label{sec:experience}
\vspace{4pt}
\begin{table*}[t]
	\centering
	\setlength{\tabcolsep}{1mm}
	{
		\small
		\begin{tabular}{c|l|cc:cc|cc:cc:cc}
			\Xhline{1pt}
			& & \multicolumn{2}{c:}{\textbf{LLaMA-3 8B}} & \multicolumn{2}{c|}{\textbf{LLaMA-3 70B}} 
			& \multicolumn{2}{c:}{\textbf{LLaMA-2 7B}} & \multicolumn{2}{c:}{\textbf{LLaMA-2 13B}} 
			& \multicolumn{2}{c}{\textbf{LLaMA-2 70B}} \\
			\cdashline{3-12}
			\textbf{\#Bits} & \textbf{Method} 
			& 0-shot$^9$ & Wiki & 0-shot$^9$ & Wiki 
			& 0-shot$^9$ & Wiki & 0-shot$^9$ & Wiki 
			& 0-shot$^9$ & Wiki \\
			\textbf{W-A-KV} & & Avg.($\uparrow$) & ($\downarrow$) & Avg.($\uparrow$) & ($\downarrow$) 
			& Avg.($\uparrow$) & ($\downarrow$) & Avg.($\uparrow$) & ($\downarrow$) 
			& Avg.($\uparrow$) & ($\downarrow$) \\
			\hdashline
			16-16-16 & FloatingPoint 
			& 68.09 & 6.14 & 73.81 & 2.86 
			& 65.21 & 5.47 & 67.61 & 4.88 
			& 71.59 & 3.32 \\
			\hdashline
			\multirow{6}{*}{4-16-16} 
			& GPTQ        & 61.03 & 7.43 & 31.45 & 9e3 & 60.86 & 9.84 & 64.71 & 5.79 & 70.96 & 3.94 \\
			& AWQ         & 67.03 & 7.36 & 68.92 & 5.92 & 63.89 & 5.83 & 66.25 & 5.07 & 70.88 & 4.03 \\
			& QuaRot      & 67.27 & 6.53 & 72.93 & 3.53 & 64.30 & 5.62 & 66.95 & 5.00 & 71.21 & 3.41 \\
			& SpinQuant   & 66.54 & 6.49 & 72.90 & \textbf{3.49} & 63.59 & \textbf{5.58} & 67.14 & 5.00 & 71.12 & 3.43 \\
			& \textcolor{gray}{OSTQuant} & \textcolor{gray}{67.80} & \textcolor{gray}{6.53} & \textcolor{gray}{73.69} & \textcolor{gray}{3.19} & \textcolor{gray}{64.37} & \textcolor{gray}{5.64}  & \textcolor{gray}{67.31} & \textcolor{gray}{4.94} & \textcolor{gray}{71.48} & \textcolor{gray}{3.41} \\
			& \textbf{\ours} & \textbf{67.36} & \textbf{6.48} & \textbf{73.25} & 3.50 & \textbf{64.34} & 5.60 & \textbf{67.27} & \textbf{4.99} & \textbf{71.25} & \textbf{3.40} \\
			\hdashline
			\multirow{4}{*}{4-4-16} 
			& QuaRot      & 61.69 & 8.02 & 65.56 & 6.35 & 61.87 & 6.05 & 65.13 & 5.35 & 69.96 & 3.78 \\
			& SpinQuant   & 64.11 & 7.28 & 66.99 & 6.10 & 57.37 & 6.78 & 63.23 & 5.24 & 70.58 & 3.68 \\
			& \textcolor{gray}{OSTQuant} & \textcolor{gray}{65.14} & \textcolor{gray}{7.24} & \textcolor{gray}{72.21} & \textcolor{gray}{3.97} & \textcolor{gray}{63.90} & \textcolor{gray}{5.60} & \textcolor{gray}{66.24} & \textcolor{gray}{5.14} & \textcolor{gray}{70.92} & \textcolor{gray}{3.57} \\
			& \textbf{\ours} & \textbf{65.74} & \textbf{7.07} & \textbf{70.71} & \textbf{5.24} & \textbf{62.84} & \textbf{5.86} & \textbf{66.71} & \textbf{5.15} & \textbf{70.82} & \textbf{3.62} \\
			\hdashline
			\multirow{8}{*}{\cellcolor{white}4-4-4} 
			& QuaRot      & 61.38 & 8.18 & 65.33 & 6.60 & 61.48 & 6.11 & 65.16 & 5.39 & 70.30 & 3.80 \\
			& SpinQuant   & 64.10 & 7.35 & 66.31 & 6.24 & 62.01 & 5.96 & 64.13 & 5.74 & \textbf{70.57} & \textbf{3.61} \\
			& Kurtail & - & 7.20 & - & 7.20 & - & 5.90 & - & 5.20 & - & - \\
			& \textcolor{gray}{OSTQuant}	& \textcolor{gray}{65.37} & \textcolor{gray}{7.29} & \textcolor{gray}{71.69} & \textcolor{gray}{4.01} & \textcolor{gray}{63.18} & \textcolor{gray}{5.91}  & \textcolor{gray}{65.41} & \textcolor{gray}{5.25}  & \textcolor{gray}{70.84} & \textcolor{gray}{3.59} \\
			& $\text{OSTQuant}^\dag$ & 65.13 & \textbf{6.80} & - & - & 62.45 & \textbf{5.38} & - & - & - & - \\
			& BASE-Q & 65.39  & 7.17 & OOM & OOM & 62.50 & 5.85 & 65.48 & 5.21 &OOM  &OOM  \\
			& \textbf{\ourse} & 64.79 & 7.21 & 70.01 & 5.57 & 62.62 & 5.93 & 66.12 & 5.22 & 70.10 & 3.84 \\
			& \textbf{\ours} & \textbf{65.57} & 7.16 & \textbf{70.21} & \textbf{5.39} & \textbf{63.16} & 5.89 & \textbf{66.33} & \textbf{5.18} & 70.35 & 3.64 \\
			\Xhline{1pt}
		\end{tabular}
	}
		\caption{Comparison of perplexity on WikiText2 and averaged accuracy across nine diverse zero-shot tasks. Results for GPTQ, AWQ, QuaRot, SpinQuant, and OSTQuant are reported from the OSTQuant paper, while BASE-Q results are based on official code (Note: 'OOM' denotes out of memory on our device). Gray OSTQuant entries use distillation and are included as a strong supervised reference. \ourse denotes the application of a complete global orthogonal rotation. $\text{OSTQuant}^\dag$ refers to OSTQuant without distillation.}
	\label{MainResults}
\end{table*}

\begin{table}[t]
	\centering
	\caption{Results on Qwen2.5 models. Quantized methods use W4A4KV4. QuaRot and SpinQuant are reproduced from their official implementations; OSTQuant on 32B is unavailable due to out-of-memory errors.}
	\label{QWENResults}
	\small
	\setlength{\tabcolsep}{2.5pt}
	\begin{tabular}{lcccc}
		\toprule
		\multirow{2}{*}{\textbf{Method}}
		& \multicolumn{2}{c}{\textbf{Qwen2.5 14B}}
		& \multicolumn{2}{c}{\textbf{Qwen2.5 32B}} \\
		\cmidrule(lr){2-3}\cmidrule(lr){4-5}
		& \textbf{0-shot$^9$ $\uparrow$} & \textbf{Wiki $\downarrow$}
		& \textbf{0-shot$^9$ $\uparrow$} & \textbf{Wiki $\downarrow$} \\
		\midrule
		FloatingPoint & 70.95 & 5.29 & 71.11 & 5.02 \\
		QuaRot        & 67.23 & 6.77 & 68.14 & 6.04 \\
		SpinQuant     & 67.29 & 6.55 & 68.51 & 5.88 \\
		\textcolor{gray}{OSTQuant}
		               & \textcolor{gray}{67.81} & \textcolor{gray}{6.37}
		               & \textcolor{gray}{OOM} & \textcolor{gray}{OOM} \\
		\textbf{\ours} & \textbf{67.65} & \textbf{6.30}
		               & \textbf{69.90} & \textbf{5.69} \\
		\bottomrule
	\end{tabular}
\end{table}

\paragraph{Models and Datasets.}
We evaluate whether activation-distribution optimization transfers across model families, scales, and evaluation metrics. The comparison covers LLaMA-2 (7B--70B)~\citep{touvron2023llama}, LLaMA-3 (8B--70B)~\citep{grattafiori2024llama}, and Qwen2.5 (14B/32B)~\citep{qwen2025qwen25technicalreport}. We report WikiText2 perplexity (PPL)~\citep{merity2016pointer} as a sensitive language-modeling metric and use nine zero-shot tasks from \texttt{lm-evaluation-harness} (version 0.4.7)~\citep{eval-harness} to check whether lower quantization error translates to task-level behavior. The zero-shot suite includes BoolQ~\citep{clark2019boolq}, HellaSwag~\citep{zellers2019hellaswag}, LAMBADA (OpenAI)~\citep{radford2019language}, OpenBookQA (OBQA)~\citep{mihaylov2018can}, PIQA~\citep{bisk2020piqa}, SIQA~\citep{sap2019socialiqa}, WinoGrande~\citep{sakaguchi2021winogrande}, ARC-Easy, and ARC-Challenge~\citep{boratko2018systematic}.
\paragraph{Baselines.} We compare with representative quantization methods that stress different parts of the design space: weight reconstruction methods GPTQ~\citep{frantar2023optq} and AWQ~\citep{lin2024awq}, rotation-based methods QuaRot~\citep{ashkboos2024quarot} and SpinQuant~\citep{liu2025spinquant}, and recent low-bit LLM quantizers Kurtail~\citep{akhondzadeh2025kurtail}, BASE-Q~\citep{he2025base}, and OSTQuant~\citep{hu2025ostquant}. This comparison is useful because \ours changes the activation distribution before quantization, whereas several baselines mainly reconstruct weights, use fixed rotations, or rely on stronger supervision; in particular, the distilled OSTQuant results are included as a strong supervised reference. Here, ``-'' denotes a result not reported in the baseline's official paper.
\paragraph{Implementation Details.}\label{details}
The calibration process uses 128 samples from WikiText-2, each with a sequence length of 2048. Activations are quantized using per-token asymmetric quantization, while weights are quantized using asymmetric per-channel quantization with GPTQ~\citep{frantar2023optq}, applying a group size of 128 for key-value matrices. During the \our phase, we optimize block-diagonal orthogonal matrices initialized with Hadamard matrices via \textit{Cayley SGD}~\citep{liefficient}.
The \AT hyperparameters are selected by grid search on the calibration set, and the final search space and chosen values are reported in Appendix~\ref{sec:appx_implementation} and Appendix~\ref{app:appx_moreablation}.
We report two variants to separate accuracy and deployment considerations. \ours uses block-diagonal rotations to reduce transformation overhead, while \ourse uses a full global orthogonal rotation similar to SpinQuant. More implementation details are provided in Appendix~\ref{sec:appx_implementation}.
\subsection{Main Results}
\paragraph{Quantization Performance.}
Table~\ref{MainResults} shows that the value of activation-distribution optimization becomes visible when activations are quantized. In the weight-only 4-16-16 setting, \ours preserves $98.7\%$--$99.5\%$ of the floating-point zero-shot accuracy across the evaluated LLaMA models, but the gap among strong rotation-based methods is relatively small. This pattern is informative: when activations remain in high precision, reshaping them is not the dominant bottleneck. The setting mainly verifies that the learned rotation does not damage the weight-only quantization path.

Once activations are quantized, the table reveals different behavior. Under 4-4-16, \ours improves the average zero-shot accuracy over SpinQuant by $2.91$ points across the five LLaMA settings, with larger gains on LLaMA-2 7B ($+5.47$) and LLaMA-3 70B ($+3.72$). Under W4A4KV4, where weights, activations, and KV cache are all quantized to 4 bits, \ours preserves $96.9\%$ of floating-point accuracy on average and improves over SpinQuant by $1.70$ points. These results support the main design intuition of the paper: fixed or generic rotations are often sufficient to avoid catastrophic outliers, but low-bit activation quantization benefits from learning a distribution that uses the available quantization levels more evenly.
This trend transfers to Qwen2.5 in Table~\ref{QWENResults}. Among W4A4KV4 methods, \ours achieves the lowest perplexity on both 14B and 32B; its 14B zero-shot accuracy is within $0.16$ points of the distillation-based OSTQuant reference, while its 32B result outperforms the available rotation baselines.

The comparison with OSTQuant also clarifies the boundary of the method. \ours outperforms the distillation-based OSTQuant baseline on LLaMA-3 8B and LLaMA-2 13B, while OSTQuant remains competitive on several 70B entries. This mixed pattern is useful rather than merely negative: it suggests that distribution shaping can recover a large part of the activation-quantization loss without full-precision supervision, but supervision and scale-specific calibration may still help in some large-model regimes. Full per-task results are reported in Appendix~\ref{sec:appx_results}.
\paragraph{Speedup and Memory Savings.}
\begin{table}[ht]
	\vspace{-2pt}
	\centering
	\setlength{\tabcolsep}{0.8mm}{
		\small
		\begin{tabular}{l|ccc|ccc}
			\Xhline{1pt}
			\multirow{2}{*}{\textbf{Method}} & \multicolumn{3}{c|}{\textbf{Prefill Speedup }}                       & \multicolumn{3}{c}{\textbf{Memory Saving}}                         \\ \cline{2-7} 
			& \textbf{2048}                 & \textbf{4096}                 & \textbf{8192}                  & \textbf{2048}                 & \textbf{4096}                 & \textbf{8192}                 \\ \hline

			2-70B-\ours                   & 2.46                 & 2.11                 & 1.97                  & 2.91                 & 2.59                 & 2.22                 \\
			2-70B-\ourse                  & 2.61                 & 2.26                 & 2.10                  & 3.25                 & 2.84                 & 2.36                 \\  \hline
						3-8B-\ours                    & 1.57                 & 1.43                 & 1.28                  & 2.46                 & 2.12                 & 1.86    \\
			3-8B-\ourse                   & 1.74                 & 1.55                 & 1.42                  & 2.78                 & 2.37                 & 2.00                 \\
			\Xhline{1pt}
		\end{tabular}
	}
	\caption{Speedup and memory savings factors for LLaMA models of different sizes and sequence lengths, comparing 4-bit quantized implementations to FP16.}
	\vspace{-1pt}
\label{tab:speed}
\end{table}
We evaluate inference efficiency using the W4A4 kernel from~\citet{ashkboos2024quarot}. Under this kernel, QuaRot, SpinQuant, and \ourse share the same inference computation graph and therefore the same theoretical speedup and memory savings; we omit duplicate QuaRot and SpinQuant rows from Table~\ref{tab:speed}. The table reports prefill speedup and decoding memory savings relative to FP16 on a single Transformer block, using batch size 4 on an NVIDIA RTX 4090. On LLaMA-2 70B with sequence length 2048, \ourse achieves a $2.61\times$ prefill speedup and $3.25\times$ memory saving, while \ours achieves a $2.46\times$ prefill speedup and $2.91\times$ memory saving. The difference between \ours and \ourse exposes the main deployment trade-off: full rotations can be slightly faster in this implementation path, whereas block-diagonal rotations provide a more flexible per-layer optimization structure.

\begin{figure}[htp]
	\vspace{-1pt}
	\centering \includegraphics[width=1\linewidth]{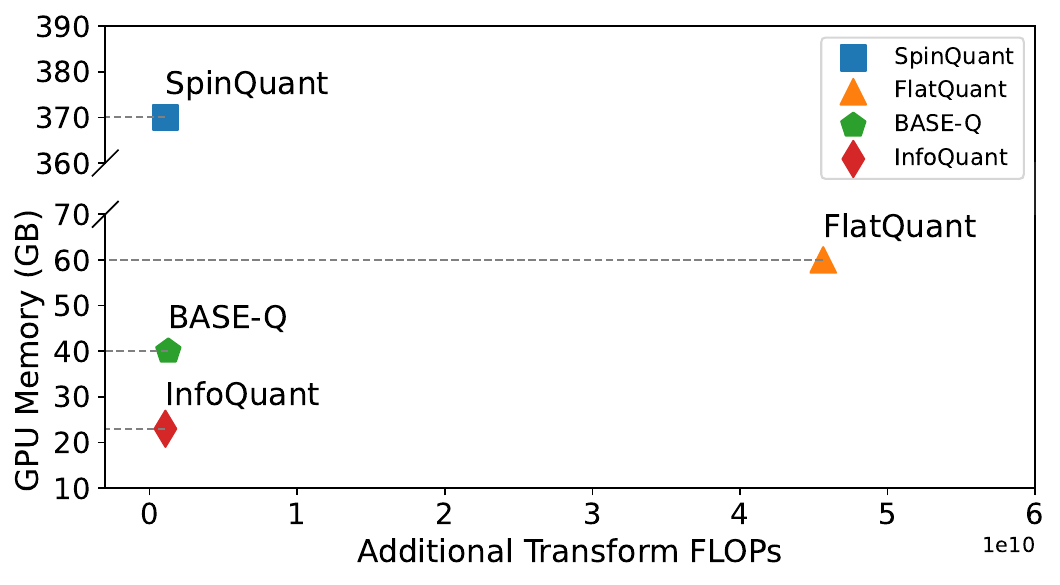}
	\caption{Comparison of memory overhead in quantization and transform FLOPs during inference across different methods. During inference, FlatQuant performs transformations on three activation values online and dynamically, while other methods apply the fast Hadamard transform only to the KV-cache efficiently.}
	\label{fig:speed}
	\vspace{-1pt}
\end{figure}

We further compare quantization-time memory overhead and inference-time transformation FLOPs on LLaMA-3 70B. As shown in Figure~\ref{fig:speed}, \ourse requires less than 24GB of GPU memory during quantization and introduces low additional transformation cost. The broader insight is that activation shaping should not be evaluated only by accuracy: a transformation that must be repeatedly applied online can erase part of the benefit of low-bit inference. FlatQuant, for example, performs multiple dynamic activation transformations during inference, which increases transform FLOPs and depends on specialized kernels. \ours instead keeps the rotation-based deployment path lightweight, making the method more compatible with existing low-bit kernels and consumer-grade quantization hardware.
\subsection{Ablation Study} \label{sec:abla}
\paragraph{Module-wise Impact.}
Table~\ref{ablation_table} isolates how each component contributes to \ours under the W4A4KV4 setting. Replacing a fixed Hadamard rotation with the learned \our rotation reduces WikiText2 perplexity from $10.90$ to $8.84$ on LLaMA-3 8B and from $8.99$ to $7.03$ on LLaMA-2 7B. This is the largest single change in the ablation, which suggests that the core gain comes from learning where activation energy should be redistributed, not from simply adding more calibration stages.

The later components produce smaller but more diagnostic changes. Adding GPTQ further reduces perplexity to $7.53$ and $6.01$, indicating that activation-oriented rotation remains compatible with standard weight reconstruction. \AT gives a modest improvement, which is consistent with its role as a signal reweighting mechanism rather than a separate quantizer. LAC then gives the best perplexity on both models ($7.16$ and $5.89$). The ordering of these gains is important: first reshape the distribution, then reconstruct weights, then refine which calibration tokens and clipping ranges deserve attention.
\begin{table}[htbp]
	\renewcommand\arraystretch{1}
	\centering
	\setlength{\tabcolsep}{1mm}{
	\small
		\begin{tabular}{ccccc|cc
			}
			\Xhline{1pt}
			\multirow{2}{*}{\textbf{Hadamard}} & \multirow{2}{*}{\textbf{\our}} & \multirow{2}{*}{\textbf{GPTQ}}     & \multirow{2}{*}{\textbf{\AT}}     & \multirow{2}{*}{\textbf{LAC}}      & \multicolumn{2}{c}{\textbf{WikiText2}} \\ \cline{6-7} 
			&                             &                           &                           &                           & \textbf{3-8B}       & \textbf{2-7B}       \\ \hline
			\checkmark                    &                             &                           &                           &                           & 10.9            & 8.99            \\
			& \checkmark                 &                           &                           &                           & 8.84            & 7.03            \\
			& \checkmark                 & \checkmark               &                           &                           & 7.53            & 6.01            \\
			& \checkmark                 & \checkmark               & \checkmark               &                           & 7.42            & 5.97            \\
			& \checkmark                 & \checkmark               & \checkmark               & \checkmark               & \textbf{7.16}   & \textbf{5.89}   \\ \Xhline{1pt}
		\end{tabular}
	}
	\caption{Component ablation of \ours under the W4A4KV4 configuration. Lower WikiText2 perplexity indicates better quantization quality.}
\label{ablation_table}
\end{table}

\paragraph{More Ablations.}
Additional ablations in Appendix~\ref{app:appx_moreablation} study the weighting factor $\gamma$, the temperature $T$, initialization robustness, clipping-ratio sensitivity, and the block size of block-diagonal orthogonal matrices. These experiments provide practical guidance for using \ours: objective reweighting and temperature control the stability of peak suppression, while block size and clipping range determine how much distribution-shaping flexibility can be traded for lower inference cost and easier calibration.

\vspace{2pt}
\section{Conclusion}
\vspace{2pt}
This work studies low-bit LLM quantization from the perspective of quantizer-facing activation distribution design. By analyzing the distributional error introduced by quantization, we show that quantization-friendly activations should jointly have a smaller numerical range and sufficient dispersion within that range. Guided by this principle, \ours reshapes activations with \ourone, emphasizes informative calibration tokens with \AT, and refines the quantization interval with learnable activation clipping. Our results suggest that activation quantization should be evaluated by both outlier reduction and effective use of available low-bit levels. A natural next step is to extend this analysis beyond bell-shaped activation assumptions and develop transformations for broader distributional regimes.

\section*{Limitations}\label{sec:limit}
\vspace{-0.5mm}
The findings of this paper should be interpreted within the scope of the evaluated settings. We validate \ours on a limited set of model families, tasks, and quantization configurations, and do not claim that the same gains will automatically transfer to substantially different architectures, activation regimes, or deployment scenarios without additional study. Our method is motivated by activation patterns that have been widely observed in prior studies and are also present in the models evaluated in this paper. While these observations are sufficient to support the improvements reported here, broader validation would still be useful to determine how consistently the same behavior holds across other model families and quantization settings.
From a practical perspective, the method still relies on calibration data and implementation choices such as transformation and clipping settings. Although the approach is intended for practical post-training quantization, its effectiveness may therefore vary when calibration conditions differ substantially from those used in evaluation.

\section*{Acknowledgments}
This work was supported in part by The National Nature Science Foundation of China (Grant Nos. 62303406 and 62432014) and in part by Ant Group.
\bibliography{custom}

@article{touvron2023llama,
  title={Llama 2: Open foundation and fine-tuned chat models},
  author={Touvron, Hugo and Martin, Louis and Stone, Kevin and Albert, Peter and Almahairi, Amjad and Babaei, Yasmine and Bashlykov, Nikolay and Batra, Soumya and Bhargava, Prajjwal and Bhosale, Shruti and others},
  journal={arXiv preprint arXiv:2307.09288},
  year={2023}
}

@article{grattafiori2024llama,
  title={The llama 3 herd of models},
  author={Grattafiori, Aaron and Dubey, Abhimanyu and Jauhri, Abhinav and Pandey, Abhinav and Kadian, Abhishek and Al-Dahle, Ahmad and Letman, Aiesha and Mathur, Akhil and Schelten, Alan and Vaughan, Alex and others},
  journal={arXiv preprint arXiv:2407.21783},
  year={2024}
}

@article{ashkboos2024quarot,
  title={Quarot: Outlier-free 4-bit inference in rotated llms},
  author={Ashkboos, Saleh and Mohtashami, Amirkeivan and Croci, Maximilian and Li, Bo and Cameron, Pashmina and Jaggi, Martin and Alistarh, Dan and Hoefler, Torsten and Hensman, James},
  journal={Advances in Neural Information Processing Systems},
  volume={37},
  pages={100213--100240},
  year={2024}
}

@inproceedings{MaLZLX0W0J24,
  author={Yuexiao Ma and Huixia Li and Xiawu Zheng and Feng Ling and Xuefeng Xiao and Rui Wang and Shilei Wen and Fei Chao and Rongrong Ji},
  title={AffineQuant: Affine Transformation Quantization for Large Language Models},
  year={2024},
  cdate={1704067200000},
  url={https://openreview.net/forum?id=of2rhALq8l},
  booktitle={ICLR},
}

@misc{sun2025flatquantflatnessmattersllm,
      title={FlatQuant: Flatness Matters for LLM Quantization}, 
      author={Yuxuan Sun and Ruikang Liu and Haoli Bai and Han Bao and Kang Zhao and Yuening Li and Jiaxin Hu and Xianzhi Yu and Lu Hou and Chun Yuan and Xin Jiang and Wulong Liu and Jun Yao},
      year={2025},
      eprint={2410.09426},
      archivePrefix={arXiv},
      primaryClass={cs.CL},
      url={https://arxiv.org/abs/2410.09426}, 
}

@inproceedings{
liu2025spinquant,
title={SpinQuant: {LLM} Quantization with Learned Rotations},
author={Zechun Liu and Changsheng Zhao and Igor Fedorov and Bilge Soran and Dhruv Choudhary and Raghuraman Krishnamoorthi and Vikas Chandra and Yuandong Tian and Tijmen Blankevoort},
booktitle={The Thirteenth International Conference on Learning Representations},
year={2025},
url={https://openreview.net/forum?id=ogO6DGE6FZ}
}

@inproceedings{
hu2025ostquant,
title={{OSTQ}uant: Refining Large Language Model Quantization with Orthogonal and Scaling Transformations for Better Distribution Fitting},
author={Xing Hu and Yuan Cheng and Dawei Yang and Zhixuan Chen and Zukang Xu and JiangyongYu and XUCHEN and Zhihang Yuan and Zhe jiang and Sifan Zhou},
booktitle={The Thirteenth International Conference on Learning Representations},
year={2025},
url={https://openreview.net/forum?id=rAcgDBdKnP}
}

@inproceedings{frantar2023optq,
  title={OPTQ: Accurate post-training quantization for generative pre-trained transformers},
  author={Frantar, Elias and Ashkboos, Saleh and Hoefler, Torsten and Alistarh, Dan-Adrian},
  booktitle={11th International Conference on Learning Representations},
  year={2023}
}

@article{merity2016pointer,
  title={Pointer sentinel mixture models},
  author={Merity, Stephen and Xiong, Caiming and Bradbury, James and Socher, Richard},
  journal={arXiv preprint arXiv:1609.07843},
  year={2016}
}

@misc{eval-harness,
  author       = {Gao, Leo and Tow, Jonathan and Abbasi, Baber and Biderman, Stella and Black, Sid and DiPofi, Anthony and Foster, Charles and Golding, Laurence and Hsu, Jeffrey and Le Noac'h, Alain and Li, Haonan and McDonell, Kyle and Muennighoff, Niklas and Ociepa, Chris and Phang, Jason and Reynolds, Laria and Schoelkopf, Hailey and Skowron, Aviya and Sutawika, Lintang and Tang, Eric and Thite, Anish and Wang, Ben and Wang, Kevin and Zou, Andy},
  title        = {A framework for few-shot language model evaluation},
  month        = 07,
  year         = 2024,
  publisher    = {Zenodo},
  version      = {v0.4.3},
  doi          = {10.5281/zenodo.12608602},
  url          = {https://zenodo.org/records/12608602}
}

@article{clark2019boolq,
  title={BoolQ: Exploring the surprising difficulty of natural yes/no questions},
  author={Clark, Christopher and Lee, Kenton and Chang, Ming-Wei and Kwiatkowski, Tom and Collins, Michael and Toutanova, Kristina},
  journal={arXiv preprint arXiv:1905.10044},
  year={2019}
}

@article{sap2019socialiqa,
  title={Socialiqa: Commonsense reasoning about social interactions},
  author={Sap, Maarten and Rashkin, Hannah and Chen, Derek and LeBras, Ronan and Choi, Yejin},
  journal={arXiv preprint arXiv:1904.09728},
  year={2019}
}

@article{zellers2019hellaswag,
  title={Hellaswag: Can a machine really finish your sentence?},
  author={Zellers, Rowan and Holtzman, Ari and Bisk, Yonatan and Farhadi, Ali and Choi, Yejin},
  journal={arXiv preprint arXiv:1905.07830},
  year={2019}
}

@article{sakaguchi2021winogrande,
  title={Winogrande: An adversarial winograd schema challenge at scale},
  author={Sakaguchi, Keisuke and Bras, Ronan Le and Bhagavatula, Chandra and Choi, Yejin},
  journal={Communications of the ACM},
  year={2021}
}

@article{mihaylov2018can,
  title={Can a suit of armor conduct electricity? a new dataset for open book question answering},
  author={Mihaylov, Todor and Clark, Peter and Khot, Tushar and Sabharwal, Ashish},
  journal={arXiv preprint arXiv:1809.02789},
  year={2018}
}

@inproceedings{bisk2020piqa,
  title={Piqa: Reasoning about physical commonsense in natural language},
  author={Bisk, Yonatan and Zellers, Rowan and Gao, Jianfeng and Choi, Yejin and others},
  booktitle={Proceedings of the AAAI conference on artificial intelligence},
  volume={34},
  pages={7432--7439},
  year={2020}
}

@article{boratko2018systematic,
  title={A systematic classification of knowledge, reasoning, and context within the ARC dataset},
  author={Boratko, Michael and Padigela, Harshit and Mikkilineni, Divyendra and Yuvraj, Pritish and Das, Rajarshi and McCallum, Andrew and Chang, Maria and Fokoue-Nkoutche, Achille and Kapanipathi, Pavan and Mattei, Nicholas and others},
  journal={arXiv preprint arXiv:1806.00358},
  year={2018}
}

@article{radford2019language,
  title={Language models are unsupervised multitask learners},
  author={Radford, Alec and Wu, Jeffrey and Child, Rewon and Luan, David and Amodei, Dario and Sutskever, Ilya and others},
  journal={OpenAI blog},
  volume={1},
  number={8},
  pages={9},
  year={2019}
}

@inproceedings{xiao2023smoothquant,
  title={Smoothquant: Accurate and efficient post-training quantization for large language models},
  author={Xiao, Guangxuan and Lin, Ji and Seznec, Mickael and Wu, Hao and Demouth, Julien and Han, Song},
  booktitle={International Conference on Machine Learning},
  pages={38087--38099},
  year={2023},
  organization={PMLR}
}

@inproceedings{liefficient,
  title={Efficient Riemannian Optimization on the Stiefel Manifold via the Cayley Transform},
  author={Li, Jun and Li, Fuxin and Todorovic, Sinisa},
  booktitle={International Conference on Learning Representations},
  year={2020},
}

@inproceedings{zhao2024lrquant,
  title={LRQuant: Learnable and Robust Post-Training Quantization for Large Language Models},
  author={Zhao, Jiaqi and Zhang, Miao and Zeng, Chao and Wang, Ming and Liu, Xuebo and Nie, Liqiang},
  booktitle={Proceedings of the 62nd Annual Meeting of the Association for Computational Linguistics (Volume 1: Long Papers)},
  pages={2240--2255},
  year={2024}
}

@inproceedings{
akhondzadeh2025kurtail,
title={{KURTAIL} : {KURTOSIS}-{BASED} {LLM} {QUANTIZATION}},
author={Mohammad Sadegh Akhondzadeh and Aleksandar Bojchevski and Evangelos Eleftheriou and Martino Dazzi},
booktitle={Sparsity in LLMs (SLLM): Deep Dive into Mixture of Experts, Quantization, Hardware, and Inference},
year={2025},
url={https://openreview.net/forum?id=GYVIWuazp5}
}

@inproceedings{
tseng2024quip,
title={Qu{IP}\${\textbackslash}\#\$: Even Better {LLM} Quantization with Hadamard Incoherence and Lattice Codebooks},
author={Albert Tseng and Jerry Chee and Qingyao Sun and Volodymyr Kuleshov and Christopher De Sa},
booktitle={Forty-first International Conference on Machine Learning},
year={2024},
url={https://openreview.net/forum?id=9BrydUVcoe}
}

@article{chee2023quip,
  title={Quip: 2-bit quantization of large language models with guarantees},
  author={Chee, Jerry and Cai, Yaohui and Kuleshov, Volodymyr and De Sa, Christopher M},
  journal={Advances in Neural Information Processing Systems},
  volume={36},
  pages={4396--4429},
  year={2023}
}

@article{lin2024duquant,
  title={Duquant: Distributing outliers via dual transformation makes stronger quantized llms},
  author={Lin, Haokun and Xu, Haobo and Wu, Yichen and Cui, Jingzhi and Zhang, Yingtao and Mou, Linzhan and Song, Linqi and Sun, Zhenan and Wei, Ying},
  journal={Advances in Neural Information Processing Systems},
  volume={37},
  pages={87766--87800},
  year={2024}
}

@inproceedings{
liu2025trainingfree,
title={Training-Free Activation Sparsity in Large Language Models},
author={James Liu and Pragaash Ponnusamy and Tianle Cai and Han Guo and Yoon Kim and Ben Athiwaratkun},
booktitle={The Thirteenth International Conference on Learning Representations},
year={2025},
url={https://openreview.net/forum?id=dGVZwyq5tV}
}

@article{lin2024awq,
  title={Awq: Activation-aware weight quantization for on-device llm compression and acceleration},
  author={Lin, Ji and Tang, Jiaming and Tang, Haotian and Yang, Shang and Chen, Wei-Ming and Wang, Wei-Chen and Xiao, Guangxuan and Dang, Xingyu and Gan, Chuang and Han, Song},
  journal={Proceedings of Machine Learning and Systems},
  volume={6},
  pages={87--100},
  year={2024}
}

@inproceedings{
shao2024omniquant,
title={OmniQuant: Omnidirectionally Calibrated Quantization for Large Language Models},
author={Wenqi Shao and Mengzhao Chen and Zhaoyang Zhang and Peng Xu and Lirui Zhao and Zhiqian Li and Kaipeng Zhang and Peng Gao and Yu Qiao and Ping Luo},
booktitle={The Twelfth International Conference on Learning Representations},
year={2024},
url={https://openreview.net/forum?id=8Wuvhh0LYW}
}

@inproceedings{
yi2025rotated,
title={Rotated Runtime Smooth: Training-Free Activation Smoother for accurate {INT}4 inference},
author={Ke Yi and Zengke Liu and jianwei zhang and Chengyuan Li and Tong Zhang and Junyang Lin and Jingren Zhou},
booktitle={The Thirteenth International Conference on Learning Representations},
year={2025},
url={https://openreview.net/forum?id=WG7GzGx3G9}
}

@inproceedings{lee2024owq,
  title={Owq: Outlier-aware weight quantization for efficient fine-tuning and inference of large language models},
  author={Lee, Changhun and Jin, Jungyu and Kim, Taesu and Kim, Hyungjun and Park, Eunhyeok},
  booktitle={Proceedings of the AAAI Conference on Artificial Intelligence},
  volume={38},
  pages={13355--13364},
  year={2024}
}

@inproceedings{
ashkboos2024slicegpt,
title={Slice{GPT}: Compress Large Language Models by Deleting Rows and Columns},
author={Saleh Ashkboos and Maximilian L. Croci and Marcelo Gennari do Nascimento and Torsten Hoefler and James Hensman},
booktitle={The Twelfth International Conference on Learning Representations},
year={2024},
url={https://openreview.net/forum?id=vXxardq6db}
}

@inproceedings{wei2023outlier,
  title={Outlier suppression+: Accurate quantization of large language models by equivalent and effective shifting and scaling},
  author={Wei, Xiuying and Zhang, Yunchen and Li, Yuhang and Zhang, Xiangguo and Gong, Ruihao and Guo, Jinyang and Liu, Xianglong},
  booktitle={Proceedings of the 2023 Conference on Empirical Methods in Natural Language Processing},
  pages={1648--1665},
  year={2023}
}

@misc{chen2025prefixquanteliminatingoutliersprefixed,
      title={PrefixQuant: Eliminating Outliers by Prefixed Tokens for Large Language Models Quantization}, 
      author={Mengzhao Chen and Yi Liu and Jiahao Wang and Yi Bin and Wenqi Shao and Ping Luo},
      year={2025},
      eprint={2410.05265},
      archivePrefix={arXiv},
      primaryClass={cs.LG},
      url={https://arxiv.org/abs/2410.05265}, 
}

@inproceedings{liu2024intactkv,
  title={IntactKV: Improving Large Language Model Quantization by Keeping Pivot Tokens Intact},
  author={Liu, Ruikang and Bai, Haoli and Lin, Haokun and Li, Yuening and Gao, Han and Xu, Zhengzhuo and Hou, Lu and Yao, Jun and Yuan, Chun},
  booktitle={ACL (Findings)},
  year={2024}
}

@misc{qwen2025qwen25technicalreport,
      title={Qwen2.5 Technical Report}, 
      author={Qwen and : and An Yang and Baosong Yang and Beichen Zhang and Binyuan Hui and Bo Zheng and Bowen Yu and Chengyuan Li and Dayiheng Liu and Fei Huang and Haoran Wei and Huan Lin and Jian Yang and Jianhong Tu and Jianwei Zhang and Jianxin Yang and Jiaxi Yang and Jingren Zhou and Junyang Lin and Kai Dang and Keming Lu and Keqin Bao and Kexin Yang and Le Yu and Mei Li and Mingfeng Xue and Pei Zhang and Qin Zhu and Rui Men and Runji Lin and Tianhao Li and Tianyi Tang and Tingyu Xia and Xingzhang Ren and Xuancheng Ren and Yang Fan and Yang Su and Yichang Zhang and Yu Wan and Yuqiong Liu and Zeyu Cui and Zhenru Zhang and Zihan Qiu},
      year={2025},
      eprint={2412.15115},
      archivePrefix={arXiv},
      primaryClass={cs.CL},
      url={https://arxiv.org/abs/2412.15115}, 
}

@article{he2025base,
  title={BASE-Q: Bias and Asymmetric Scaling Enhanced Rotational Quantization for Large Language Models},
  author={He, Liulu and Zheng, Shenli and Sun, Karwei and Liu, Yijiang and Zhao, Yufei and Tan, Chongkang and Yang, Huanrui and Du, Yuan and Du, Li},
  journal={arXiv preprint arXiv:2506.15689},
  year={2025}
}

\appendix
\section{Theoretical Details}
\label{sec:appx_proofs}

\subsection{Smoothed KL Approximation and Error Bound}
\begin{theorem}
\label{theorem1}
Let $X\sim P$ be a centered symmetric bell-shaped activation variable with standard deviation $\sigma$, and let
\[
Q_{s,c}(x)=\operatorname{clip}\!\left(s\left\lfloor \frac{x}{s}\right\rceil,\,-c,\,c\right)
\]
be a finite-bit uniform round-to-nearest quantizer with step size $s$, clipping scale $c=Ms$, and $M=2^{B-1}-1$. Define the quantization centroids and cells by
\[
\begin{aligned}
q_i &= is,\qquad i=-M,\ldots,M,\\
I_i &= \{x\mid Q_{s,c}(x)=q_i\},
\end{aligned}
\]
and the smoothed quantized density
\[
\begin{aligned}
Q_\theta^{(c)}(x)&=\sum_{i=-M}^{M} p_i\,\delta(x-q_i;\theta),\\
p_i&=\int_{I_i}P(x)\,dx,
\end{aligned}
\]
where $\delta(x;\theta)=\frac{1}{2\theta}\exp(-|x|/\theta)$ and $\theta\ll s$. Then the smoothed KL objective admits the approximation
\[
\begin{aligned}
D_{\mathrm{KL}}(P\|Q_\theta^{(c)})
&\approx -H(P) + H(\{p_i\}) + \log(2\theta) \\
&\quad + \frac{1}{\theta}\mathcal{E}_{\mathrm{clip}},
\end{aligned}
\]
where $\mathcal{E}_{\mathrm{clip}}=\mathbb{E}|X-Q_{s,c}(X)|$ is the expected absolute error of the clamped quantizer. Moreover, the normalized clipped error
\[
\mathcal{E}_{\mathrm{clip}}'=\frac{1}{\sigma}\mathbb{E}|X-Q_{s,c}(X)|
\]
satisfies
\[
\begin{aligned}
\mathcal{E}_{\mathrm{clip}}' &\le \frac{\lambda}{2}+\tau_P(\kappa),\\
\lambda &= \frac{s}{\sigma},\qquad
\kappa = \frac{c}{\sigma},
\end{aligned}
\]
where $\tau_P(\kappa)=\mathbb{E}[(|Y|-\kappa)_+]$ for $Y=X/\sigma$. With fixed bit-width $B$, this bound can be written as
\[
\begin{aligned}
F_{P,B}(\kappa) &= \frac{\kappa}{2M}+\tau_P(\kappa),\\
\kappa &= \frac{1}{b_n}.
\end{aligned}
\]
For Gaussian and Laplace activations, $\tau_P$ has the closed forms given below, and $F_{P,B}(\kappa)$ is increasing in $\kappa$ once the clipping tail is sufficiently controlled. Equivalently, in this tail-controlled regime, a smaller clipped range relative to the activation spread and a larger max-normalized dispersion $b_n=\sigma/c$ tighten the bound.
\end{theorem}

\subsection{Proof of Theorem 1}
For the finite-bit clamped quantizer, the interior cells are
\[
\begin{aligned}
I_i&=\left[\left(i-\frac{1}{2}\right)s,\left(i+\frac{1}{2}\right)s\right),\\
&\qquad |i|<M,
\end{aligned}
\]
while the boundary cells absorb the clipped tails:
\[
\begin{aligned}
I_M &= \left[c-\frac{s}{2},\infty\right),\\
I_{-M} &= \left(-\infty,-c+\frac{s}{2}\right].
\end{aligned}
\]
For $x\in I_i$, the quantizer outputs the centroid $q_i=is$. Under the kernel-separation assumption $\theta\ll s$, neighboring kernels contribute negligibly around each cell, and we may approximate
\[
\begin{aligned}
Q_\theta^{(c)}(x)
&\approx p_i\,\delta(x-q_i;\theta)\\
&= \frac{p_i}{2\theta}\exp\!\left(-\frac{|x-q_i|}{\theta}\right),\qquad x\in I_i.
\end{aligned}
\]
Substituting this expression into the KL divergence gives
\begin{align}
	D_{\mathrm{KL}}(P\|Q_\theta^{(c)})
	&= \sum_{i=-M}^{M}\int_{I_i} P(x) \nonumber \\
	&\quad \times \log\!\left(\frac{P(x)}{Q_\theta^{(c)}(x)}\right)dx \nonumber \\
	&\approx \sum_{i=-M}^{M}\int_{I_i} P(x)\log P(x)\,dx \nonumber \\
	&\quad - \sum_{i=-M}^{M}\log(p_i)\int_{I_i}P(x)\,dx \nonumber \\
	&\quad + \frac{1}{\theta}\sum_{i=-M}^{M}\int_{I_i}|x-q_i|P(x)\,dx \nonumber \\
	&\quad + \log(2\theta).
\end{align}
Using
\[
\begin{aligned}
\sum_i\int_{I_i}P(x)\log P(x)\,dx &= -H(P),\\
\sum_i \log(p_i)\int_{I_i}P(x)\,dx &= \sum_i p_i\log p_i\\
&= -H(\{p_i\}),
\end{aligned}
\]
we obtain
\begin{align}
	D_{\mathrm{KL}}(P\|Q_\theta^{(c)})
	&\approx -H(P)+H(\{p_i\})+\log(2\theta) \nonumber \\
	&\quad + \frac{1}{\theta}\mathcal{E}_{\mathrm{clip}},
\end{align}
where
\[
\begin{aligned}
\mathcal{E}_{\mathrm{clip}}
&=\sum_{i=-M}^{M}\int_{I_i}|x-q_i|P(x)\,dx\\
&=\mathbb{E}|X-Q_{s,c}(X)|.
\end{aligned}
\]

To bound the practical error under clamped quantization, define the clipping projection $\Pi_c(x)=\operatorname{clip}(x,-c,c)$. Then
\begin{align}
	|x-Q_{s,c}(x)|
	&\le |x-\Pi_c(x)| \nonumber \\
	&\quad + |\Pi_c(x)-Q_{s,c}(x)| \nonumber \\
	&\le (|x|-c)_+ + \frac{s}{2}.
\end{align}
The second term is the standard in-range round-to-nearest error, while the first term is the clipping-tail error. Taking expectations and dividing by $\sigma$ gives
\begin{align}
	\mathcal{E}_{\mathrm{clip}}'
	&:= \frac{1}{\sigma}\mathbb{E}|X-Q_{s,c}(X)| \nonumber \\
	&\le \frac{s}{2\sigma} + \frac{1}{\sigma}\mathbb{E}\!\left[(|X|-c)_+\right].
\end{align}
With the normalized variable $Y=X/\sigma$, $\lambda=s/\sigma$, and $\kappa=c/\sigma$, this becomes
\begin{align}
	\mathcal{E}_{\mathrm{clip}}'
	&\le \frac{\lambda}{2} + \tau_P(\kappa), \nonumber \\
	\tau_P(\kappa)
	&:= \mathbb{E}\!\left[(|Y|-\kappa)_+\right].
\end{align}
Because $c=Ms$ for a fixed $B$-bit quantizer, we have $\kappa=M\lambda$ and therefore
\[
F_{P,B}(\kappa)=\frac{\kappa}{2M}+\tau_P(\kappa).
\]

\paragraph{Gaussian activations.}
If $X\sim\mathcal{N}(0,\sigma^2)$, then $Y\sim\mathcal{N}(0,1)$ with density $\phi(y)=\frac{1}{\sqrt{2\pi}}e^{-y^2/2}$ and survival function $\bar{\Phi}(y)=1-\Phi(y)$. The clipping-tail term is
\begin{align}
	\tau_G(\kappa)
	&= 2\int_{\kappa}^{\infty}(y-\kappa)\phi(y)\,dy \nonumber \\
	&= 2\big(\phi(\kappa)-\kappa\bar{\Phi}(\kappa)\big).
\end{align}
Hence
\[
\begin{aligned}
F_{G,B}(\kappa)
&=\frac{\kappa}{2M}+2\big(\phi(\kappa)-\kappa\bar{\Phi}(\kappa)\big),
\end{aligned}
\]
and
\[
F_{G,B}'(\kappa)=\frac{1}{2M}-2\bar{\Phi}(\kappa).
\]
Therefore $F_{G,B}$ is increasing whenever
\[
\kappa \ge \kappa_G^\star := \Phi^{-1}\!\left(1-\frac{1}{4M}\right).
\]
In this tail-controlled regime, decreasing $\kappa=c/\sigma$ reduces the bound.

\paragraph{Laplace activations.}
If $X\sim\mathrm{Laplace}(0,b)$, then $\sigma=\sqrt{2}\,b$. We keep the same normalized error definition as above,
\[
\mathcal{E}_{\mathrm{clip}}'=\frac{1}{\sigma}\mathbb{E}|X-Q_{s,c}(X)|,
\]
so the Laplace case is parameterized by the scale $b$ only through the identity $\sigma=\sqrt{2}b$. With the normalized variable $Y=X/\sigma$, the density becomes
\[
f_Y(y)=\frac{1}{\sqrt{2}}e^{-\sqrt{2}|y|}.
\]
The clipping-tail term becomes
\begin{align}
	\tau_L(\kappa)
	&= 2\int_{\kappa}^{\infty}(y-\kappa)\frac{1}{\sqrt{2}}e^{-\sqrt{2}y}\,dy \nonumber \\
	&= \frac{1}{\sqrt{2}}e^{-\sqrt{2}\kappa}.
\end{align}
Hence
\[
\begin{aligned}
F_{L,B}(\kappa)&=\frac{\kappa}{2M}+\frac{1}{\sqrt{2}}e^{-\sqrt{2}\kappa},\\
F_{L,B}'(\kappa)&=\frac{1}{2M}-e^{-\sqrt{2}\kappa}.
\end{aligned}
\]
Therefore $F_{L,B}$ is increasing whenever
\[
\kappa \ge \kappa_L^\star := \frac{\log(2M)}{\sqrt{2}}.
\]
Again, in this regime, decreasing $\kappa$ tightens the bound. Equivalently, one may express the same condition in terms of $b$ via $\kappa=c/(\sqrt{2}b)$, but the normalized error itself remains defined with respect to $\sigma$.

Finally, the normalized dispersion metric used in the main text is
\[
b_n=\frac{\sigma}{c}=\frac{1}{\kappa},
\qquad
\lambda=\frac{s}{\sigma}=\frac{s/c}{\sigma/c}=\frac{\bar{s}}{b_n}.
\]
Therefore, for a fixed bit-width and a controlled clipping tail, a smaller clipped range relative to the activation spread and a larger max-normalized dispersion $b_n$ both correspond to a smaller $\kappa$ and a tighter error bound.

\subsection{Effect of \our on Normalized Dispersion}
\label{prof:omm}
Let $\mathbf{t}\in\mathbb{R}^{d}$ be an activation token and $\mathbf{A}$ an orthogonal matrix. After rotation and centering as in Eq.~\eqref{eq:ommin}, $\mathbf{t}'=\mathbf{t}\mathbf{A}-\mathbb{E}[\mathbf{t}\mathbf{A}]$. Orthogonal rotation preserves the Euclidean norm, while centering can only remove the mean component:
\begin{align}
	\|\mathbf{t}'\|_2 \le \|\mathbf{t}\mathbf{A}\|_2 = \|\mathbf{t}\|_2
\end{align}
For a centered token, the normalized dispersion is
\begin{align}
	b_n(\mathbf{t}') =
	\sqrt{\frac{1}{d}}
	\frac{\|\mathbf{t}'\|_2}{\|\mathbf{t}'\|_\infty}.
\end{align}
\begin{align}
	\frac{b_n(\mathbf{t}')}{b_n(\mathbf{t})}
	=
	\frac{\|\mathbf{t}'\|_2}{\|\mathbf{t}\|_2}
	\cdot
	\frac{\|\mathbf{t}\|_\infty}{\|\mathbf{t}'\|_\infty}.
\end{align}
Thus, when \our suppresses the peak while preserving most centered-token energy, the reduction in $\|\mathbf{t}'\|_\infty$ dominates the change in $\|\mathbf{t}'\|_2$ and increases $b_n$. This is a conditional mechanism rather than a universal guarantee.

\section{Implementation and Efficiency}
\label{sec:appx_implementation}
\begin{figure}[htbp]
	\centering
	\includegraphics[width=\linewidth]{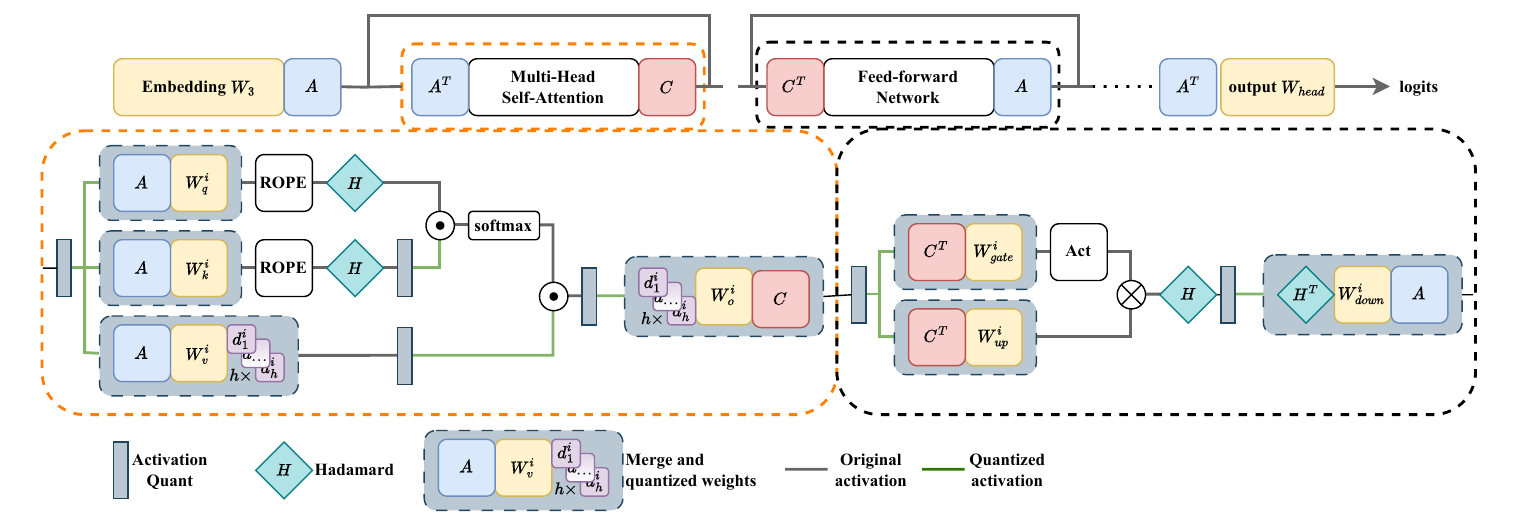}
	\caption{Global-rotation computation graph of \ourse.}
	\label{app:group}
\end{figure}

\begin{figure}[htbp]
	\centering
	\includegraphics[width=\linewidth]{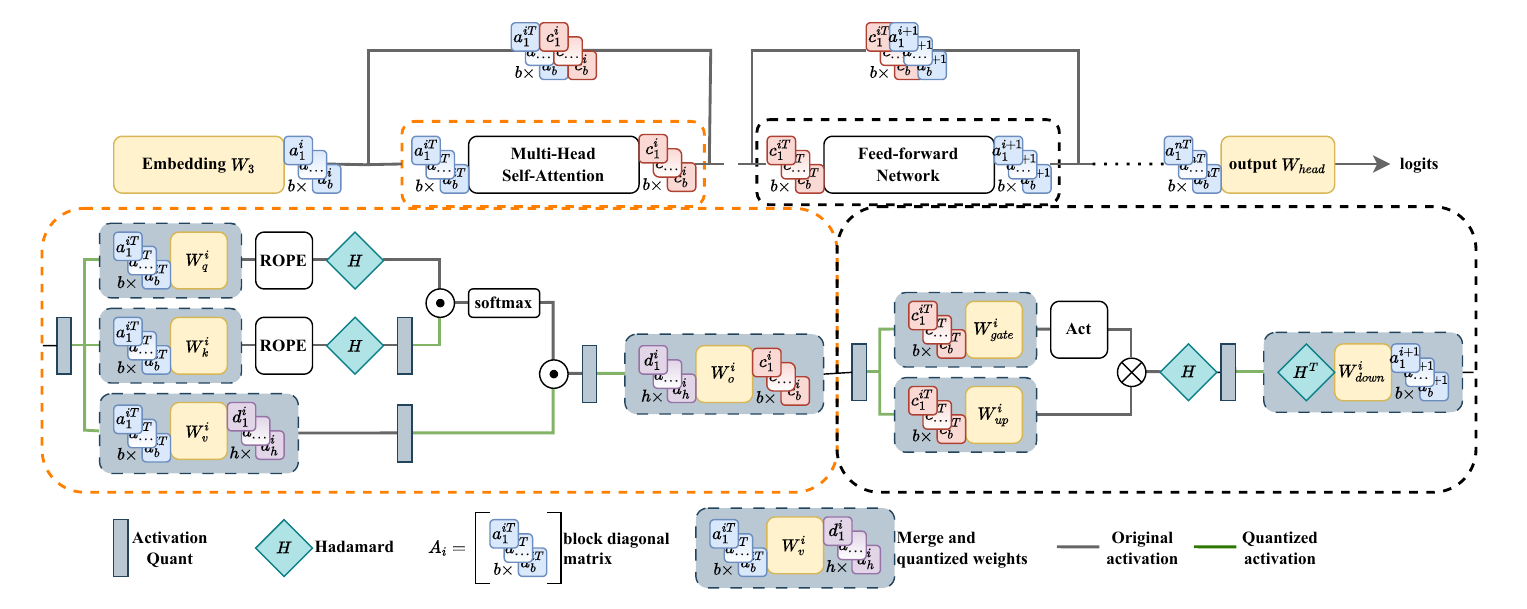}
	\caption{Layer-specific block-diagonal rotation graph of \ours.}
	\label{app:group_res}
\end{figure}

\paragraph{Additional Setup.}
We partition the block-diagonal rotations into two blocks for 7B--13B models and four blocks for the 70B model. \our uses $T=2$, batch size $4$, and an initial learning rate of $2$ linearly decayed over 15 epochs. For \AT, we use $m=10$ calibration samples and select $\delta=0.02$, $\tau=0.4$, and $\gamma=30$ by grid search. LAC initializes the clipping ratio at $0.95$ within $[0.5,1]$ and uses AdamW with batch size $4$ and an initial learning rate of $0.05$ cosine-annealed over 5 epochs.
\paragraph{Computational Graph.}
\ours optimizes layer-specific activation rotations, which require an additional matrix multiplication at residual connections (Figure~\ref{app:group_res}). Block-diagonal matrices reduce this overhead while retaining per-layer flexibility. In contrast, \ourse uses a global rotation (Figure~\ref{app:group}), avoids the residual-connection multiplication, and has a computation graph equivalent to SpinQuant~\citep{liu2025spinquant}.

\paragraph{Runtime and Memory Overhead.}\label{app:runtime}
\begin{table}[htbp]
	\centering

	\setlength{\tabcolsep}{1mm}{
	\small
	\begin{tabular}{lccccc}
		\toprule
		\multirow{2}{*}{\textbf{Model}} & \multicolumn{2}{c}{\textbf{\our}} & \multicolumn{2}{c}{\textbf{LAC}} \\
		\cmidrule(lr){2-3} \cmidrule(lr){4-5}
		& \textbf{Time} & \textbf{Memory} & \textbf{Time} & \textbf{Memory} \\
		\midrule
		LLaMA-2 7B   & $\sim$6.0  & $\sim$6  & $\sim$3.0  & $\sim$10 \\
		LLaMA-2 13B  & $\sim$7.5  & $\sim$7  & $\sim$2.0  & $\sim$12 \\
		LLaMA-2 70B  & $\sim$13.0 & $\sim$11 & $\sim$3.5  & $\sim$23 \\
		\bottomrule
	\end{tabular}
	}
	\caption{Per-layer optimal time (min) and GPU memory usage (GB) on NVIDIA RTX 4090.}
\label{tab:runtime_memory}
\end{table}
Table~\ref{tab:runtime_memory} reports the per-layer optimization runtime and memory consumption for LLaMA-2 7B, 13B, and 70B on an NVIDIA RTX 4090. The 70B configuration remains within 24GB of GPU memory.
\section{Ablations and Robustness}
\label{app:appx_moreablation}
\subsection{Optimization Choices}
\paragraph{Influence of Weighting Factor $\gamma$.}
\begin{table}[htbp]
	\vspace{-1pt}
	\renewcommand\arraystretch{1}
	\centering
	\setlength{\tabcolsep}{1mm}{
		\small
		\begin{tabular}{c|c}
			\Xhline{1pt}
			\textbf{Weighting factor} $\gamma$ & \textbf{Wikitext2 PPL} \\ \hline
			1 & 7.03 \\
			30 & \textbf{6.89} \\
			50 & 6.97 \\
			80 & 6.93 \\
			$+\infty$ & 7.02 \\ \Xhline{1pt}
		\end{tabular}
	}
	\caption{Ablation study of weighting factor $\gamma$ on WikiText2 PPL using RTN quantization under W4A4KV4 configuration for LLaMA-2 7B.}
	\vspace{-1pt}
	\label{tab:ablation_gamma}
\end{table}
Table~\ref{tab:ablation_gamma} shows that $\gamma=30$ best balances emphasis on outlier tokens with retention of normal calibration tokens.
\paragraph{Influence of Temperature $T$ in \our.}
The temperature controls how selectively \our weights high-magnitude coordinates. Table~\ref{tab:ablation_T} shows that $T=2$ performs best, whereas overly sharp or flat weighting increases perplexity.
\begin{table}[htbp]
	\vspace{-2pt}
	\renewcommand\arraystretch{1}
	\centering
	\setlength{\tabcolsep}{1mm}{
	\small
		\begin{tabular}{c|c}
	\Xhline{1pt}
	\textbf{Temperature} $T$ & \textbf{Wikitext2 PPL} \\ \hline
	0.3 & 7.20 \\ 
	1   & 7.11 \\ 
	2   & \textbf{6.89} \\ 
	4   & 7.04 \\
	8   & 7.22\\
	\Xhline{1pt}
\end{tabular}
	}
	\caption{Effect of the \our temperature $T$ on WikiText2 perplexity using RTN quantization under the W4A4KV4 configuration for LLaMA-2 7B.}
	\label{tab:ablation_T}
\end{table}
\vspace{-8pt}

\subsection{Rotation Design}
\paragraph{Block Sizes of Block-diagonal Orthogonal Matrices.}
Table~\ref{tab:block} shows that increasing the number of blocks degrades perplexity by restricting cross-block redistribution, consistent with the inter-block shift observed by~\citet{lin2024duquant}. We therefore use an intermediate block size to balance distribution shaping and inference cost.
\begin{table}[htbp]
	\centering
	\caption{Impact of block number in block-diagonal orthogonal matrices on quantization performance under W4A4KV4 configuration with GPTQ.}
	\label{tab:block}
	\resizebox{0.38\textwidth}{!}{
\begin{tabular}{c|cc}
				\Xhline{1pt}
	\textbf{Block number} & \textbf{LLaMA-3 8B} & \textbf{LLaMA-2 13B} \\ \hline
	1            &     \textbf{7.13}       &      \textbf{5.16}       \\
	2            &     7.18       &      5.18      \\
	4            &     7.17       &      5.32       \\
	8            &     7.39       &      5.37       \\ 	\Xhline{1pt}
\end{tabular}
}
\end{table}
\paragraph{Robustness of Initialization $\mathbf{A}$.}
Table~\ref{tab:init} compares Hadamard initialization with random orthogonal matrices obtained by QR decomposition. Their nearly identical perplexities on LLaMA-2 13B and LLaMA-3 8B indicate that \our is insensitive to the initial orthogonal basis.
\begin{table}[htbp]
	\centering
	\caption{Impact of initialization of \our under W4A4KV4 configuration with GPTQ.}
	\label{tab:init}
		\resizebox{0.4\textwidth}{!}{
	\begin{tabular}{c|cc}
				\Xhline{1pt}
	\textbf{Model}    & \textbf{Initialization} & \textbf{WikiText2 PPL} \\ \hline
	\multirow{2}{*}{LLaMA-3 8B}  & Hadamard  & 7.18  \\
									& Random &  7.18  \\ \hline
	\multirow{2}{*}{LLaMA-2 13B} &  Hadamard &   5.18  \\
	&  Random    &    5.19    \\ 		\Xhline{1pt}
\end{tabular}
}
\end{table}
\subsection{Calibration Robustness}
We evaluate \ours on Qwen2.5-7B-Instruct under W4A4KV4 using C4 or WikiText2 calibration sets of different sizes, with five random seeds per configuration.
\begin{table}[htbp]
	\centering
	\caption{Perplexity under different calibration sources and set sizes. Results are reported as mean $\pm$ standard deviation over five random seeds.}
	\label{tab:calibration_robustness}
	\small
	\setlength{\tabcolsep}{3pt}
	\begin{tabular}{lccc}
		\toprule
		\textbf{Calib. source} & \textbf{\# Samples} & \multicolumn{2}{c}{\textbf{PPL $\downarrow$}} \\
		\cmidrule(lr){3-4}
		& & \textbf{WikiText2} & \textbf{C4} \\
		\midrule
		C4        & 128 & $13.66 \pm 0.06$ & $17.91 \pm 0.06$ \\
		C4        & 256 & $13.23 \pm 0.09$ & $17.73 \pm 0.03$ \\
		C4        & 512 & $13.05 \pm 0.17$ & $17.45 \pm 0.02$ \\
		\midrule
		WikiText2 & 128 & $13.48 \pm 0.14$ & $17.94 \pm 0.04$ \\
		WikiText2 & 256 & $13.17 \pm 0.08$ & $17.78 \pm 0.05$ \\
		WikiText2 & 512 & $12.72 \pm 0.07$ & $17.60 \pm 0.09$ \\
		\bottomrule
	\end{tabular}
\end{table}
The standard deviation is at most $0.17$ on WikiText2 and $0.09$ on C4, showing stable performance across calibration sources, set sizes, and seeds.
\subsection{Clipping Sensitivity}
Table~\ref{tab:ablation_clip} evaluates the clipping ratios $(\alpha,\beta)$ after \our; moderate clipping improves perplexity, while aggressive clipping degrades it.
\begin{table}[htbp]
	\renewcommand\arraystretch{1}
	\centering
	\caption{WikiText2 perplexity of LLaMA-2 7B under different clipping ratios after \our, using RTN quantization with W4A4KV4.}
	\label{tab:ablation_clip}
	\setlength{\arrayrulewidth}{0.5pt}
	\resizebox{0.3\textwidth}{!}{
		\begin{tabular}{c|c}
			\Xhline{1pt}
				\textbf{clip ratio $\alpha,\beta$} & \textbf{Wikitext2 PPL} \\ \hline
			1   & 7.21 \\ 
			0.95   & 7.06 \\ 
			0.9 & 6.98 \\ 
			0.85   & 7.00 \\
			0.8   & 8.02 \\
			\Xhline{1pt}
		\end{tabular}
	}
\end{table}
\section{Full Results}
\label{sec:appx_results}
Table~\ref{tab:appendix_main_llama23} reports the complete per-task results using the same nine-task evaluation protocol as Section~\ref{sec:experience}.
\begin{table*}[t]
	\renewcommand\arraystretch{1}
	\centering
	\vspace{-3mm}
	\caption{Full \ours's results of the perplexity score on WikiText2 and averaged accuracy on all task on \textbf{LLaMA-2 \& 3}.}
	\vspace{3mm}
	\label{tab:appendix_main_llama23}
	\setlength{\tabcolsep}{0.8mm}
	{\resizebox{0.8\textwidth}{!}{
			\begin{tabular}{c|c|cccccccccc:c}
				\Xhline{1pt}
				\multirow{2}{*}{\textbf{Model}} & \textbf{\#Bits} & \textbf{ARC-c} & \textbf{ARC-e
				} & \textbf{BoolQ} & \textbf{HellaS.
				} & \textbf{Lam.
				} & \textbf{OBQA} & \textbf{PIQA
				} & \textbf{SIQA
				} & \textbf{WinoG.
				}  & \textbf{Avg.
				}  & \textbf{Wiki2} \\ 
				& W-A-KV & ($\uparrow$) & ($\uparrow$) & ($\uparrow$) & ($\uparrow$) & ($\uparrow$) & ($\uparrow$) & ($\uparrow$) & ($\uparrow$) & ($\uparrow$) & ($\uparrow$) & ($\downarrow$) \\
				\noalign{\vspace{0.2em}}\hline\noalign{\vspace{0.2em}}
				\multirow{4}{*}{2-7B} & 16-16-16  & 46.42  & 74.33  & 77.71  & 75.94  & 73.69  & 44.20  & 79.16  & 45.91  & 69.53  & 65.21  & 5.47  \\

				& \multirow{1}[0]{*}{4-16-16} & 44.62 & 73.86 & 76.67 & 75.22 & 72.87 & 43.60 & 78.02 & 45.65 & 68.43 & 64.34 & 5.60 \\

				& \multirow{1}[0]{*}{4-4-16}  & 43.34 & 71.25 & 75.44 &74.10 & 71.90 & 40.80 & 77.09 & 44.98 & 66.61 & 62.84 & 5.86 \\

				& \multirow{1}[0]{*}{4-4-4}  &43.17 & 71.59 & 75.60 & 73.92 & 72.39 & 42.20 & 77.69 & 45.14 & 66.77 & 63.16 & 5.89\\
				\noalign{\vspace{0.2em}}\hline\noalign{\vspace{0.2em}}
				\multirow{4}{*}{2-13B} & 16-16-16 & 49.15  & 77.53  & 80.58  & 79.39  & 76.62  & 45.20  & 80.63  & 47.49  & 71.90  & 67.61  & 4.88  \\

				& \multirow{1}{*}{4-16-16}  & 48.89 & 77.31 & 79.82 & 78.81 & 76.40 & 45.20 & 79.92 & 46.72 & 72.38 & 67.27 & 4.99\\

				& \multirow{1}{*}{4-4-16}  & 48.98 & 75.72 & 80.06 & 78.40 & 75.39 & 44.80 & 79.43 & 46.32 & 71.19 & 66.71 & 5.15\\

				& \multirow{1}{*}{4-4-4}  & 48.04 & 75.34 & 79.36 & 77.97 & 75.47 & 44.40 & 79.54 & 45.60 & 71.27 & 66.33 & 5.18\\
				\noalign{\vspace{0.2em}}\hline\noalign{\vspace{0.2em}}
				\multirow{4}{*}{2-70B} & 16-16-16 & 57.42  & 81.02  & 83.79  & 83.81  & 79.60  & 48.80  & 82.70  & 49.18  & 77.98  & 71.59  & 3.32  \\
				& \multirow{1}{*}{4-16-16}  & 57.25 & 80.86 & 82.96 & 83.37 & 79.66 & 48.20 & 82.92 & 48.82 & 77.27 & 71.25 & 3.40\\
								
				& \multirow{1}{*}{4-4-16} & 55.72 & 80.01 & 82.35 & 82.86 & 79.64 & 49.00 & 82.05 & 48.52 & 77.27 & 70.82 & 3.62\\
				
				& \multirow{1}{*}{4-4-4} & 55.89 & 80.05 &81.99& 82.53 & 78.92 & 47.60 & 82.15 & 48.67 & 75.37 & 70.35 & 3.64\\
				\noalign{\vspace{0.2em}}\hline\noalign{\vspace{0.2em}}
				\multirow{4}{*}{3-8B} & 16-16-16 & 53.50  & 77.74  & 81.10  & 79.18  & 75.74  & 44.80  & 80.63  & 47.08  & 73.01  & 68.09  & 6.14  \\

				& \multirow{1}[0]{*}{4-16-16} & 53.07 & 77.44 & 78.44 & 78.18 & 74.77 & 44.00 & 80.30 & 46.62 & 73.40 & 67.36 & 6.48\\

				& \multirow{1}[0]{*}{4-4-16} & 48.81 & 77.02 & 77.77 & 76.67 & 73.30 & 43.80 & 78.45 & 44.52 & 71.35 & 65.74 & 7.07\\

				& \multirow{1}[0]{*}{4-4-4}  & 50.00 & 75.98 & 75.65 & 76.28 & 72.46 & 44.60 & 78.67 & 45.50 & 71.03 & 65.57 &7.16 \\
				\noalign{\vspace{0.2em}}\hline\noalign{\vspace{0.2em}}
				\multirow{4}{*}{3-70B} & 16-16-16 & 64.42 & 85.98 & 85.14 & 84.95 & 79.47 & 48.46 & 84.39 & 50.82 & 80.66 & 73.81 & 2.86  \\
				
				& \multirow{1}{*}{4-16-16}  & 62.63 & 85.19 & 86.21 & 84.35 & 78.27 & 47.00 & 84.49 & 50.73 & 80.43 & 73.25 & 3.50\\
				
				& \multirow{1}{*}{4-4-16} & 57.34 & 81.14 & 84.92 & 82.88 & 77.94 & 45.80 & 81.66 & 48.41 & 76.32 & 70.71 & 5.24\\
				
				& \multirow{1}{*}{4-4-4}  & 56.66 & 81.31 & 83.61 & 82.24 & 76.23 & 45.60 & 82.05 & 47.65 & 76.56 & 70.21 & 5.39\\
				\Xhline{1pt}
	\end{tabular}}}
\end{table*}

\section{Use of LLMs}
\label{sec:appx_llms}
In this work, large language models (LLMs) were primarily utilized for tasks such as text refinement, offering writing suggestions, and improving the overall structure and clarity of the manuscript. It is important to note that LLMs did not contribute to the ideation or development of the methodology section. The authors guarantee that all LLM-generated content was thoroughly reviewed and edited to ensure its accuracy and coherence.
\section{Activation Distribution Visualizations}
\label{sec:appx_visual}
\begin{figure*}[p]
	\centering
	\includegraphics[width=0.8\linewidth]{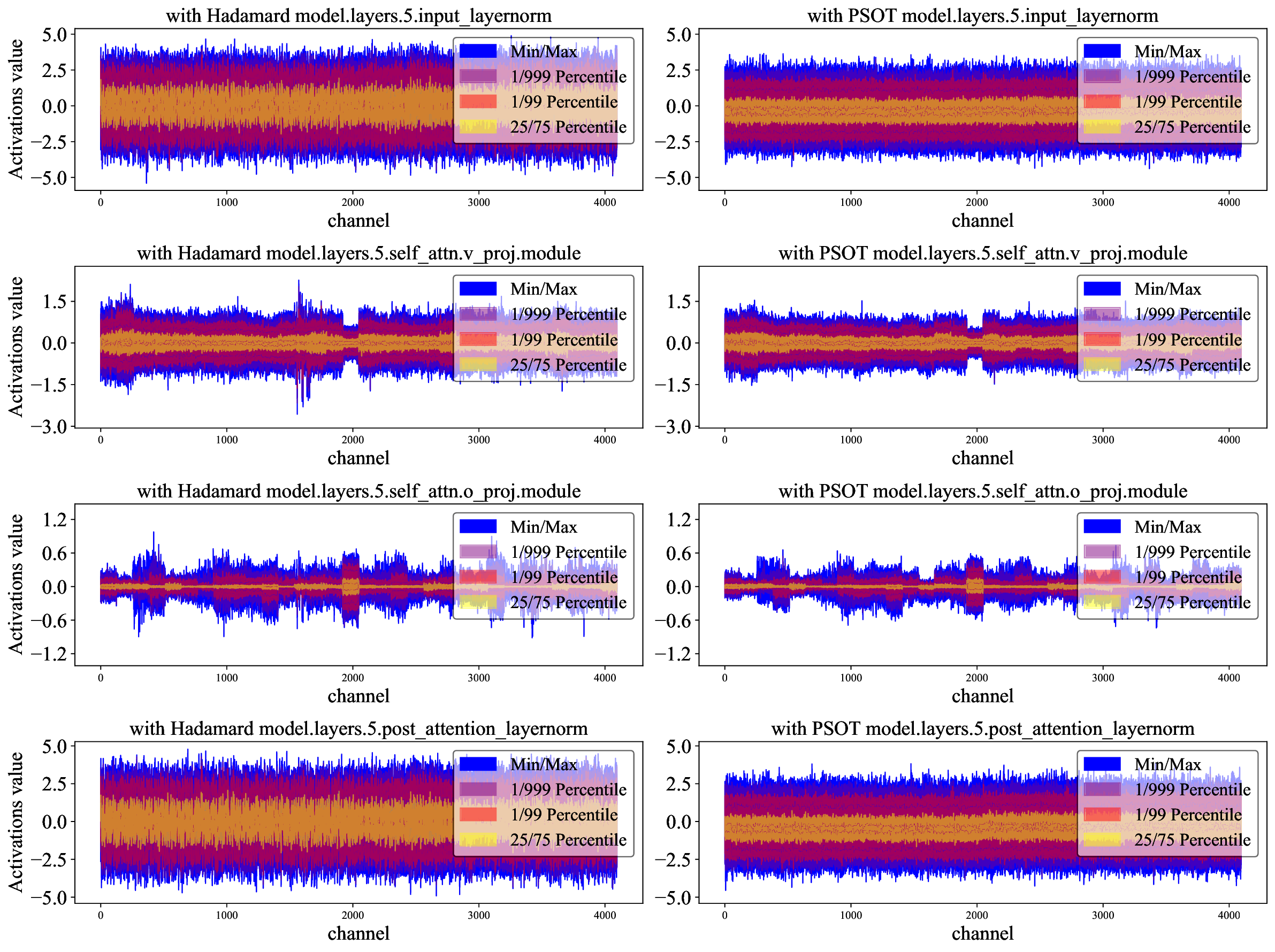}
	\caption{Activation distributions across representative LLaMA-2 7B layers and module sites after Hadamard rotation and \our.}
	\label{app:2-7b}
	\vspace{-3mm}
\end{figure*}

\begin{figure*}[p]
	\centering
	\includegraphics[width=0.8\linewidth]{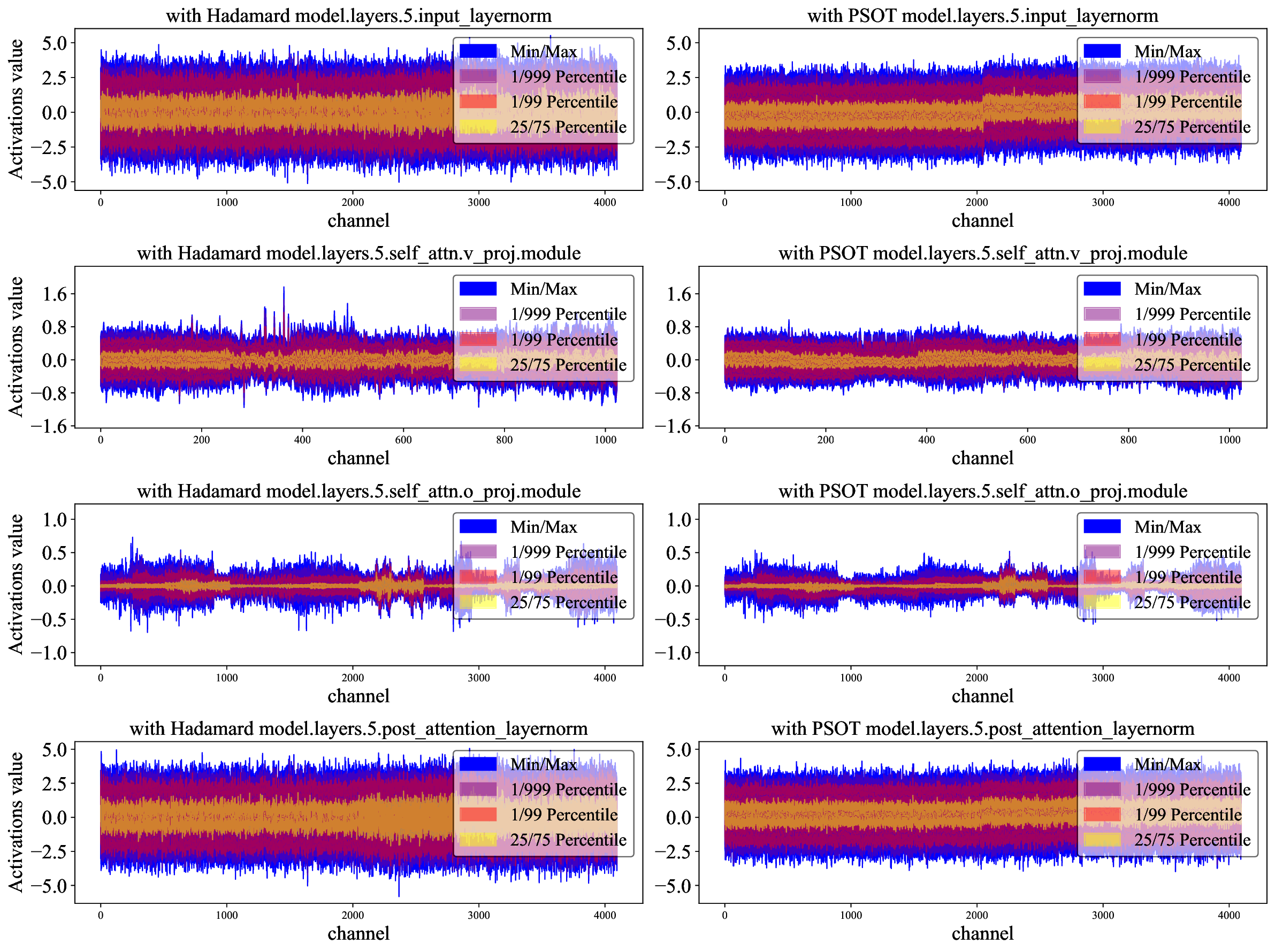}
	\caption{Activation distributions across representative LLaMA-3 8B layers and module sites after Hadamard rotation and \our.}
	\label{app:3-8b}
	\vspace{-3mm}
\end{figure*}

Figures~\ref{app:2-7b} and~\ref{app:3-8b} compare Hadamard- and \our-rotated activations across representative layers and module sites.

\end{document}